\documentclass{article}

    \usepackage[preprint]{neurips_2025}

\usepackage[utf8]{inputenc} %
\usepackage[T1]{fontenc}    %
\usepackage{hyperref}       %
\usepackage{url}            %
\usepackage{booktabs}       %
\usepackage{amsfonts}       %
\usepackage{nicefrac}       %
\usepackage{microtype}      %
\usepackage{xcolor}         %

\usepackage{microtype}
\usepackage{graphicx}
\usepackage{booktabs} %

\usepackage{hyperref}
\usepackage{url}
\usepackage{algorithm}
\usepackage{algorithmicx}
\usepackage{algpseudocode}
\usepackage{multirow}
\usepackage{multicol}
\usepackage[dvipsnames]{xcolor}
\usepackage{stfloats}
\usepackage{amsmath}
\usepackage{caption}
\usepackage{subcaption}
\usepackage{listings}

\usepackage{amsmath,amsfonts,bm}

\def\eqref#1{equation~\ref{#1}}
\def\Eqref#1{Equation~\ref{#1}}

\def\1{\bm{1}}

\def\rmD{{\mathbf{D}}}
\def\rmE{{\mathbf{E}}}

\def\vc{{\bm{c}}}

\def\vn{{\bm{n}}}

\def\vx{{\bm{x}}}
\def\vy{{\bm{y}}}
\def\vz{{\bm{z}}}

\DeclareMathAlphabet{\mathsfit}{\encodingdefault}{\sfdefault}{m}{sl}
\SetMathAlphabet{\mathsfit}{bold}{\encodingdefault}{\sfdefault}{bx}{n}

\def\gA{{\mathcal{A}}}

\def\gN{{\mathcal{N}}}

\usepackage{amsmath}
\usepackage{amssymb}
\usepackage{mathtools}
\usepackage{amsthm}

\theoremstyle{plain}

\theoremstyle{definition}

\theoremstyle{remark}

\newcommand{\modelname}{LADiBI}

\title{Blind Inverse Problem Solving Made Easy by Text-to-Image Latent Diffusion}

\author{Michail Dontas$^{1}$\thanks{Equal contribution} , Yutong He$^{1*}$, Naoki Murata$^2$, \\
\textbf{Yuki Mitsufuji$^{2,3}$, J. Zico Kolter$^{1,4}$, Ruslan Salakhutdinov$^1$} \\
Carnegie Mellon University$^1$ \quad Sony AI$^2$ \quad Sony Group Corporation$^3$ \quad Bosch Center for AI$^4$ \\
{\tt\small \{mdontas, yutonghe, zkolter, rsalakhu\}@cs.cmu.edu}\\
{\tt\small \{naoki.murata, yuhki.mitsufuji\}@sony.com}}

\begin{document}

\maketitle
\begin{abstract}
This paper considers blind inverse image restoration, the task of predicting a target image from a degraded source when the degradation (i.e. the forward operator) is unknown. Existing solutions typically rely on restrictive assumptions such as operator linearity, curated training data or narrow image distributions limiting their practicality. We introduce LADiBI, a training-free method leveraging large-scale text-to-image diffusion to solve diverse blind inverse problems with minimal assumptions. Within a Bayesian framework, LADiBI uses text prompts to jointly encode priors for both target images and operators, unlocking unprecedented flexibility compared to existing methods. Additionally, we propose a novel diffusion posterior sampling algorithm that combines strategic operator initialization with iterative refinement of image and operator parameters, eliminating the need for highly constrained operator forms. Experiments show that LADiBI effectively handles both linear and challenging nonlinear image restoration problems across various image distributions, all without task-specific assumptions or retraining.

\end{abstract}
\section{Introduction}

\label{sec:intro}
\begin{figure}[t]
    \centering
    \includegraphics[width=\textwidth]{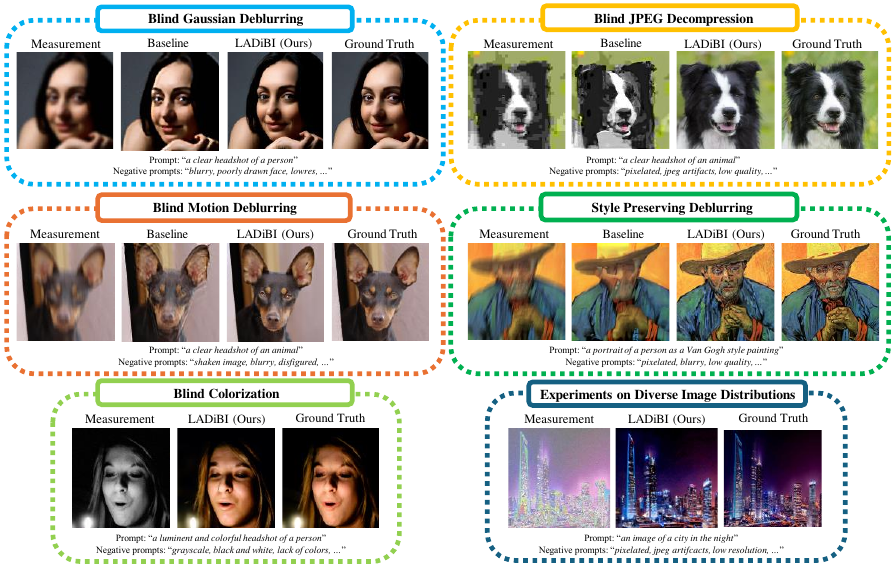}
    \caption{Our proposed \textbf{\modelname{}} is a training-free blind inverse problem solving algorithm for image restoration using large pre-trained text-to-image diffusion models. \modelname{} is applicable to a wide variety of image distribution as well as operators with minimal modeling assumptions imposed.}
    \label{fig:intro_fig}
\end{figure}
Image restoration is a critical problem in many fields such as medical imaging and computational photography, as it addresses real-world challenges including image decompression, deblurring, and super-resolution ~\citep{yuan2007, greenspan09, isaac2015}. These restoration tasks can be formulated as \emph{inverse problems}, where the goal is to recover unknown image data \(\vx\) from observed measurements \(\vy\). Formally, these problems can be expressed as \(\vy = \gA_\phi(\vx) + \vn\), where \(\gA\) is an operator representing the forward degradation process parametrized by $\phi$, and \(\vn\) is the measurement noise. Being widely applicable, this problem has attracted numerous solutions, ranging from methods with handcrafted inductive biases to deep learning, especially diffusion-based techniques~\citep{kawar2022denoising, song2022pseudoinverse, ugd, freedom, he2024manifold}.

However, most existing research focuses on the settings where the operator \(\gA_\phi\) is \emph{known}. In practice, the operator is often \emph{unknown}, leading to what are termed \emph{blind inverse problems} that present significant challenges due to their ill-posed nature. Current methods attempt to address this problem through several restrictive strategies: (1) introducing hand-crafted inductive biases via explicit formulas or specialized neural architectures~\citep{pandcp,panl0,selfdeblur}, (2) simplifying operators with linear assumptions~\citep{blinddps,murata2023gibbsddrm}, (3) constraining target distributions to narrow image classes~\citep{chihaoui2024blind,laroche2024fastem}, or (4) training task-specific models on curated measurement, operator or image datasets~\citep{mprnet,deblurganv2}. While effective in specific scenarios, these approaches suffer from limited flexibility and significant deployment barriers due to costly training and laborious hyperparameter tuning. This raises a fundamental question: can we develop a more generalizable algorithm capable of handling diverse degradation operators and image distributions without additional training or data collection?

To tackle this challenge, we reframe the problem via Bayesian inference, where we sample from the posterior $p(\vx, \gA_\phi|\vy)\propto p(\vy|\vx,\gA_\phi)p(\vx,\gA_\phi)$. This formulation naturally decomposes the seemingly intractable problem into two simpler sub-problems: estimating the prior $p(\vx,\gA_\phi)$ and sampling according to the measurement likelihood $p(\vy|\vx,\gA_\phi)$. This decomposition is particularly well-suited for diffusion-based frameworks, where we can leverage pre-trained models as the prior and guide the sampling process via measurement constraints. However, while pre-trained models typically exist for image, adequate priors or training data for the operators are generally unavailable. Thus, existing methods further factorize the joint prior into independent distributions and impose simplifying assumptions such as linearity on the operator, which severely restrict their generality and flexibility.

Instead, our approach is motivated by the key observation that, common restoration tasks can be intuitively described through natural language (e.g., ``high-definition, clear image'' for targets, ``blurry, low-quality'' for measurements). Moreover, large pre-trained text-to-image diffusion models already encapsulate rich distributions of both targets and measurements. Although this can, in some way, seem to render the approach less of a true ``blind'' solution, in practice we note that \emph{all} methods for blind inverse problems require \emph{some} assumptions over the space of transformations, and using English text to encode the joint is an extremely flexible and easy-to-use mechanism.
Based on these insights, we propose a simple yet powerful method: using classifier-free guidance~\citep{ho2021classifierfree}, we can approximate the joint prior's score across diverse images and operators using a single pre-trained text-to-image diffusion model, dramatically broadening the flexibility of existing frameworks. Our approach is particularly advantageous in blind settings, since the appropriate prompts can often be easily inferred directly from the measurements, and textual descriptions enable us to co-encode unknown degradations and desired outputs without training or handcrafted priors.

In addition to estimating the prior score, effectively sampling from the posterior distribution is also crucial for ensuring that restored images satisfy measurement constraints. In blind inverse problems, achieving this requires accurately estimating both data and operator parameters, ideally with generalized parameter classes such as neural networks to improve flexibility. However, reliably initializing these highly unconstrained operator parameters can be challenging. To address this, we propose a novel co-optimization diffusion posterior sampling algorithm specifically tailored for blind inverse problems. Our method begins with a new initialization scheme that leverages pseudo-supervision signals derived from multiple lower-quality target image approximations generated by fast posterior diffusion sampling. We then iteratively refine both operator parameters and data estimates through an alternating optimization procedure integrated within the diffusion sampling process. This strategy eliminates restrictive assumptions about operator forms, thereby enabling nonlinear blind inverse problem solving with highly flexible operator parametrizations.

Combining the text-conditioned prior with our effective posterior sampling, we introduce \textbf{L}anguage-\textbf{A}ssisted \textbf{Di}ffusion for \textbf{B}lind \textbf{I}nverse problems (\textbf{\modelname{}}), a training-free method that leverages large-scale text-to-image diffusion models to solve a broad range of blind image restoration problems with minimal assumptions. \modelname{} can be directly applied across diverse data distributions and allows for easy specification of task-specific assumptions through simple prompting.
Algorithm~\ref{alg:main_alg} and Figure~\ref{fig:alg1_fig} provide an overview of \modelname{}, which can be easily adapted from the standard inference algorithm used in popular text-to-image diffusion models. Unlike existing methods, \modelname{} requires no model retraining or reselection for different data distributions or operator functions. Instead, all prior parameterization is encoded directly in the prompt, which users can adjust as needed. Notably, we do not assume linearity of the operator, making \modelname{}, to the best of our knowledge, the most generalizable approach to blind inverse problem solving in image restoration.

We evaluate \modelname{} against state-of-the-art baselines on a range blind image restoration tasks, including linear problem (e.g. motion and Gaussian deblurring) and nonlinear problem(e.g. JPEG decompression), across various image distributions, as illustrated in Figure~\ref{fig:intro_fig}. In the linear setting, our method matches the performance of the state-of-the-art approaches while requiring significantly fewer assumptions. In the nonlinear setting, \modelname{} is the only method tested that can successfully perform JPEG decompression without any prior information of the task (such as the compression algorithm, quantization table or quantization factors), relying solely on observations of the compressed images.

\section{Background \& Related Works}
\label{sec:background}

\begin{table}[t]
\centering
\caption{A conceptual comparison between our proposed \modelname{} and the existing literature.}
\label{tab:conceptual_comparison}
\resizebox{\textwidth}{!}{
\begin{tabular}{@{}cccccc@{}}
\toprule
Method Family                        & Method        & Prior Type & Diverse Image Prior & Training-free & Flexible Operator \\ \midrule
                                     & Pan-$\ell_0$~\citep{panl0}   & Inductive bias     & {\color{red}\texttimes}                                                  & {\color{Green}\checkmark}                           & {\color{red}\texttimes}                                                \\
\multirow{-2}{*}{Optimization-based} & Pan-DCP~\citep{pandcp}   & Inductive bias    & {\color{red}\texttimes}                          & {\color{Green}\checkmark}                           & {\color{red}\texttimes}                                                \\ \midrule
Self-supervised                         & SelfDeblur~\citep{selfdeblur}    & Inductive bias & {\color{red}\texttimes}                                                  & {\color{Green}\checkmark}                                                   & {\color{red}\texttimes}                        \\ \midrule
                                     & MPRNet~\citep{mprnet}      & Discriminative  & {\color{red}\texttimes}                                                 & {\color{red}\texttimes}                                                   & {\color{Green}\checkmark}                        \\
\multirow{-2}{*}{Supervised}         & DeblurGANv2~\citep{deblurganv2}  & GAN    & {\color{red}\texttimes}                                                  & {\color{red}\texttimes}                                                   & {\color{Green}\checkmark}                        \\ \midrule
                                     & BlindDPS~\citep{blinddps}  & Pixel diffusion    & {\color{red}\texttimes}                                                  & {\color{red}\texttimes}                                                   & {\color{red}\texttimes}                                                \\
                                     & BIRD~\citep{chihaoui2024blind} & Pixel diffusion    & {\color{red}\texttimes}                                                  & {\color{Green}\checkmark}                           & {\color{red}\texttimes}                                                \\
                                     & GibbsDDRM~\citep{murata2023gibbsddrm} & Pixel diffusion    & {\color{red}\texttimes}                                                  & {\color{Green}\checkmark}                           & {\color{red}\texttimes}                                                \\
\multirow{-4}{*}{Diffusion-based}    & \textbf{\modelname{} (Ours)} & \begin{tabular}[c]{@{}c@{}}Text-to-image\\ latent diffusion\end{tabular} & {\color{Green}\checkmark}                          & {\color{Green}\checkmark}                           & {\color{Green}\checkmark}                        \\ \bottomrule
\end{tabular}
}
\end{table}

\paragraph{Diffusion for Inverse Problem Solving}
Diffusion models, also known as score-based generative models \citep{song2021scorebased, ho2020denoising}, generate clean data samples $\vx_0$ by iteratively refining noisy samples \(\vx_t\) using a time-dependent score function \(\nabla_{\vx_{t}}\log p_{t}(\vx_{t})\). This score function is usually parametrized as a noise predictor \(\epsilon_{\theta}(\vx_t, t)\) and can be used to produce the clean data samples through iteratively denoising. Particularly, a popular sampling algorithm DDIM~\citep{ddim} adopts the update rule 
\begin{align}
    \vx_{t-1} &= \sqrt{\bar{\alpha}_{t-1}} \underbrace{\left(\frac{\vx_t - \sqrt{1-\bar{\alpha}_{t}}\epsilon_{\theta}(\vx_{t}, t)}{\sqrt{\bar{\alpha}_{t}}}\right)}_{\text{intermediate estimation of $\vx_0$, denoted as }\vx_{0|t}} + \sqrt{1-\bar{\alpha}_{t-1}-\sigma_{t}^{2}}\epsilon_{\theta}(\vx_{t}, t) + \sigma_{t}\epsilon
    \label{eq:ddim}
\end{align}
that consists of an intermediate estimation of the clean data in order to perform fast sampling.

Many efforts attempt to use unconditionally pretrained diffusion for conditional generation~\citep{song2021scorebased,sdedit,dhariwal2021diffusion}, especially inverse problem solving~\citep{dps,kawar2022denoising,song2022pseudoinverse}, without additional training. Generally, when sampling from $p(\vx|\vy)$, they decompose its score as
\begin{equation}
    \nabla_{\vx_{t}}\log p_{t}(\vx_{t}|\vy) = \nabla_{\vx_{t}}\log p_{t}(\vx_{t}) + \nabla_{\vx_{t}}\log p_{t}(\vy|\vx_{t})
    \label{eq:diffusion_posterior}
\end{equation}
Since $\nabla_{\vx_{t}}\log p_{t}(\vx_{t})$ can be obtained from an unconditionally pre-trained diffusion model, these methods usually aim at deriving an accurate approximation for $\nabla_{\vx_{t}}\log p_{t}(\vy|\vx_{t})$.

Recent work has also explored text-to-image latent diffusion models 
for inverse problem-solving ~\citep{saharia2022, balaji2023ediffitexttoimagediffusionmodels, zhang2024texttoimagediffusionmodelsgenerative}. In particular, MPGD~\citep{he2024manifold} addresses inverse problems by leveraging the manifold preserving property of the latent diffusion models. Specifically, it modifies the intermediate clean latent estimate $\vz_{0|t}$ with
\begin{equation}
    \vz_{0|t} = \vz_{0|t}  - c_t\nabla_{\vz_{0|t}} \|\vy-\gA_{\phi} (\rmD(\vz_{0|t}))\|^2_2
    \label{eq:mpgd}
\end{equation}
where $\rmD(\vz_{0|t})$ represents the decoded intermediate clean image estimation and $c_t$ is the step size hyperparameter.
The $L_2$ loss $\|\vy-\gA_{\phi} (\rmD(\vz_{0|t}))\|^2_2$ is induced by the additive Gaussian noise assumption from conventional inverse problem setting. Although these diffusion-based methods perform well on diverse data distributions and tasks, they all require the operator $\gA_{\phi}$ to be known.

\paragraph{Blind Inverse Problem} 
Blind inverse problems aim to recover unknown data \(\vx \in \mathbb{R}^d\) from measurements \(\vy \in \mathbb{R}^m\), typically modeled as:
\begin{equation}
    \vy = \gA_\phi(\vx) + \vn
    \label{eq:problem_setting}
\end{equation}
where \(\gA_\phi: \mathbb{R}^d \to \mathbb{R}^m\) is the forward degradation operator parameterized by unknown function $\phi$, and \(\vn \sim \gN(0,\sigma^2 I) \in \mathbb{R}^m\) represents measurement noise with variance $\sigma^2I$.
Blind inverse problems are more difficult due to the joint estimation of \(\vx\) and \(\gA_\phi\), and are inherently ill-posed without further assumptions. 
As a result, existing methods typically incorporate different assumptions about the priors of the target image data as well as the unknown operator.

Conventional methods use hand-crafted functions as image and operator prior constraints~\citep{tv,pandcp,panl0,sparsity,tv2}. They often obtain these functional constraints by observing certain characteristics (e.g. clear edges and sparsity) unique to distributions that are usually considered as the target image (e.g. high definition natural images) and the operator (e.g. blurring kernels). However, not only are these hand-made functions not generalizable, they also often require significant manual tuning for each individual image. 

With the rise of deep learning, neural network has become a popular choice for parameterizing priors~\citep{ulyanov2018deep,gandelsman2019double,selfdeblur,deblurganv2,mprnet}. In particular, \citet{deblurganv2, mprnet} use supervised learning on data-measurement pairs to jointly learn the image-operator distributions, and \citet{ulyanov2018deep,selfdeblur} apply self-supervised learning solely on measurements using deep convolutional networks’ inherent image priors. These methods offer significant improvement over traditional approaches. However, their learning procedures require separate data collection and model training for each image distribution and task, making them resource-intensive. Although some attempt to generalize by training on a wide range of data and measurements, our experiments show that they still struggle to adapt to unseen image or operator distributions.

Recently, inspired by advances in diffusion models for inverse problem solving, numerous efforts have incorporated the priors from pre-trained diffusion models. However, most of these methods still lack generalizability~\citep{blinddps,murata2023gibbsddrm,sanghvi2023kernel,laroche2024fastem,tu2024taming,weiminbai2025blind}. For instance, \citet{blinddps,sanghvi2023kernel} require training separate operator priors, while \citet{murata2023gibbsddrm} remains training-free but rely on the operator kernel’s SVD for feasible optimization. These approaches generally depend on linear assumptions about the operator and well-trained diffusion models tailored to specific image distributions, limiting their ability to generalize across diverse image and operator types.

Table~\ref{tab:conceptual_comparison} summarizes the conceptual difference between our method and popular approaches in current literature. By leveraging large-scale pre-trained text-to-image latent diffusion models and our new posterior sampling algorithm, our method offers the most generalizability across diverse image and operator distributions with no additional training.

\section{Method}
\label{sec:method}

In this paper, we aim to tackle the problem of blind inverse problem solving defined in~\Eqref{eq:problem_setting}. Our solution has the following desiderata: (1) \textbf{No additional training:} it should not require data collection or model training, (2) \textbf{Adaptability to diverse image priors:} the same model should apply to various image distributions, (3) \textbf{Flexible operators:} no assumptions about the operator’s functional form, such as linearity or task-specific update rules, should be necessary. To make this problem feasible, we assume access to open-sourced pre-trained models.

To tackle this problem, we first formulate it as a Bayesian inference problem where the optimal solution is to sample from the posterior
\begin{equation}
    p(\vx, \gA_\phi|y) \propto p(\vy|\vx, \gA_\phi)p(\vx, \gA_\phi).
\end{equation}
This formulation allows us to decompose this problem into two parts: obtaining  sampling to maximizes the measurement likelihood $p(\vy|\vx, \gA_\phi)$. This makes diffusion-based framework the ideal approach as its posterior sampling process naturally separates these two stages.

\begin{figure*}[t]
    \centering
    \includegraphics[width=\textwidth]{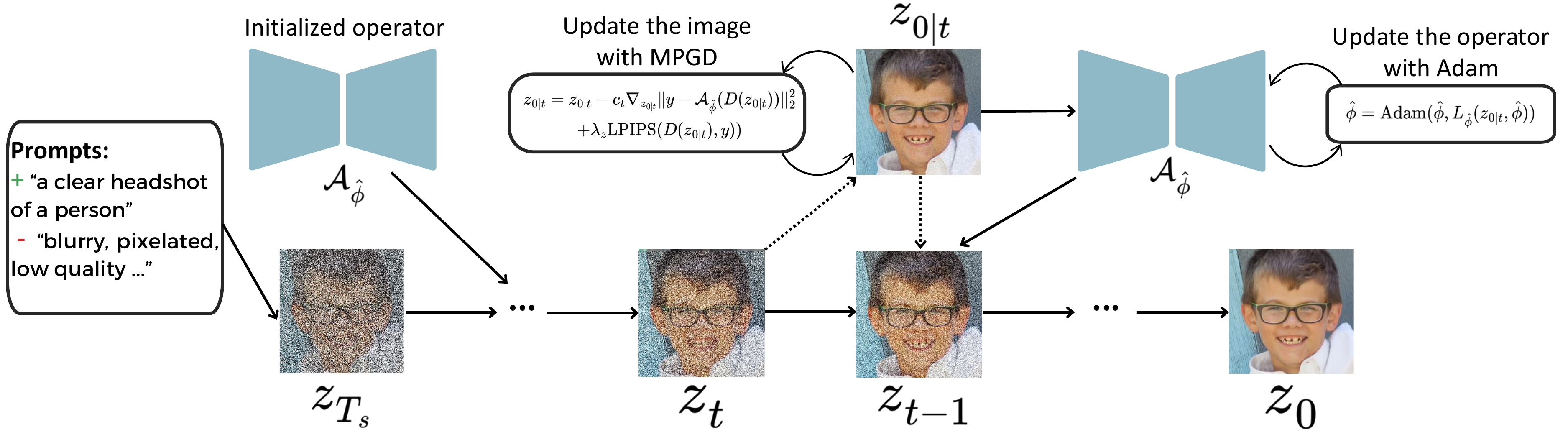}
    \caption{A schematic overview of \modelname{} (Algorithm~\ref{alg:main_alg}).}
    \label{fig:alg1_fig}
\end{figure*}

\subsection{Obtaining the prior score}
As established in~\Eqref{eq:diffusion_posterior}, diffusion-based approaches to inverse problems require access to the score of the prior distribution. For blind inverse problems, this extends to estimating the score of the joint distribution $\nabla_{\vx_t} \log p(\vx_t,\gA_\phi)$, which presents significant challenges beyond standard image priors. While strong pre-trained diffusion models usually exist for target image distributions $p(\vx)$, appropriate models or even training data for degradation operators $\gA_\phi$ are largely unavailable, making estimations of the joint distribution $p(\vx,\gA_\phi)$ particularly difficult.

Therefore, most existing diffusion-based methods further decompose the joint prior into independent distributions $p(\vx,\gA_\phi) \approx p(\vx)p(\gA_\phi)$ and impose restrictive assumptions, such as linearity constraints, on the operator~\citep{blinddps,murata2023gibbsddrm,sanghvi2023kernel,laroche2024fastem,weiminbai2025blind,tu2024taming}. To estimate the operator distribution, they either rely on simulated training data tailored to specific operator forms or perform constrained optimization under fixed operator distributions. In addition, these methods typically require different pre-trained diffusion models for different image domains. For example, deblurring human faces versus animal faces necessitate different prior models. While effective in constrained settings, their reliance on domain-specific models and strong operator assumptions fundamentally limits their generalizability, particularly violating the second and third desiderata.

Motivated by these limitations, we introduce an alternative approach inspired by the following key insights. First, many image restoration tasks can be intuitively described using natural language.
In addition, large pre-trained text-to-image diffusion models like Stable Diffusion~\citep{rombach2021highresolution} already capture rich distributions of both target images and common degradation artifacts described by these prompts.
Based on these insights, we propose to approximate the joint prior's score using large-scale pre-trained text-to-image diffusion models through classifier-free guidance~\citep{ho2021classifierfree}. In particular, we encode the desired target image characteristics using the positive prompts, and specify the unwanted degradation artifacts via a negative prompts. For example, when restoring a human face degraded by Gaussian blur, a suitable positive prompt can be ``a clear headshot of a person'', and the corresponding negative prompt could be ``blurry, low-quality image''.

Formally, denoting the positive prompt as $\vc_{+}$ and the negative prompt as $\vc_{-}$, we can approximate $\nabla_{\vx_t} \log p(\vx_t,\gA_\phi)$ as
\begin{equation}
    \nabla_{\vx_t} \log p(\vx_t,\gA_\phi) \approx \nabla_{\vx_t} \log p(\vx_t|\vc_{-}) + \gamma(\nabla_{\vx_t} \log p(\vx_t|\vc_{+}) - \nabla_{\vx_t} \log p(\vx_t|\vc_{-}))
\end{equation}
where $\gamma > 1$ is a weighting hyperparameter.
When using latent diffusion models parameterized by $\theta$, this translates to the empirical noise prediction
\begin{equation}
\small
\hat{\epsilon}_\theta(\vz_{t},t) = \epsilon_\theta(\vz_{t},t,\vc_{-}) + \gamma(\epsilon_\theta(\vz_{t},t,\vc_{+}) - \epsilon_\theta(\vz_{t},t,\vc_{-}))
\label{eq:cfg}
\end{equation}
This straightforward approximation provides access to the otherwise intractable joint prior score, and enables versatile applications across diverse image and operator distributions. By leveraging the knowledge encoded in large pre-trained text-to-image models, our method bypasses the need for task-specific training, operator restriction or model re-selection.

While this approach may appear to compromise the true "blindness", we emphasize that all blind inverse problem methods require some prior over the transformation space. Notably, textual descriptions of the target image and degradation artifacts are naturally advantageous, since users can directly infer the appropriate prompts from the measurements, which is a significantly simpler process than previous methods. These prompts provides an intuitive interface to encode diverse characteristics of both the desired outputs and the unwanted artifacts without training or handcrafted functions.

\begin{figure}[t]
    \centering
    \resizebox{0.475\linewidth}{!}{
\begin{minipage}{0.655\linewidth}
\footnotesize
\begin{algorithm}[H]
    \caption{Our algorithm \modelname{}}
    \label{alg:main_alg}
    \begin{algorithmic}[1]
    \State {\color{blue} /* Initialize latent with encoded measurement $\vy$ \& SDEdit */}
    \State $\vz_{T_s} = \sqrt{\Bar{\alpha}_{T_s}}\rmE(y) + \sqrt{1-\Bar{\alpha}_{T_s}}\epsilon_{T_s}$ for $\epsilon_{T_s} \sim\mathcal{N}(0, I)$
    \State Initialize $\hat{\phi}$ with a fixed operator prior or Algorithm~\ref{alg:operator_init}
    \For{$t=T_s,\dots,1$}
        \State {\color{blue} /* Use time-traveling for more stable results */}
        \For{$j=1,\dots,M$}
            \State {\color{blue} /* Use CFG to obtain accurate prior */}
            \State Calculate $\hat{\epsilon}_\theta(\vz_{t},t)$ with Eq.~\ref{eq:cfg}
            \State $\vz_{0|t} = \frac{1}{\sqrt{\Bar{\alpha}_t}}(\vz_{t}-\sqrt{1-\Bar{\alpha}_t}\hat{\epsilon}_\theta(\vz_{t},t))$
            \If{$j \equiv 0 \pmod 2$}
            \State {\color{blue} /* Perform MPGD with the estimated $\gA_{\hat{\phi}}$ */}
            \State Update $\vz_{0|t}$ with Eq.~\ref{eq:emprical_mpgd}
            \EndIf
            \State  $\epsilon_{t}, \epsilon'_{t} \sim\mathcal{N}(0, I)$
            \State $\vz_{t-1} = \sqrt{\Bar{\alpha}_{t-1}}\vz_{0|t} + \sqrt{1-\Bar{\alpha}_{t-1}-\sigma_{t}^{2}}\epsilon_\theta(\vz_{t}, t)+\sigma_{t}\epsilon_{t}$
            \State $\vz_{t} = \sqrt{\frac{\Bar{\alpha}_{t}}{\Bar{\alpha}_{t-1}}}\vz_{t-1} + \sqrt{1-\frac{\Bar{\alpha}_{t}}{\Bar{\alpha}_{t-1}}}\epsilon'_{t}$
            
        \EndFor
        \State {\color{blue} /* Periodically update $\hat{\phi}$ with $\vz_{0|t}$ and Eq.~\ref{eq:phi_loss} */}
        \If{$t \equiv 0 \pmod 5$}
        \For{$k=1,\dots,K$}
            \State $\hat{\phi} = \text{Adam}(\hat{\phi}, L_\phi(\vz_{0|t}, \hat{\phi}))$
        \EndFor
        \EndIf
    \EndFor
    \State \textbf{return} $\vx_0 = \rmD(\vz_0)$
    \end{algorithmic}
\end{algorithm}
\end{minipage}
}
\hfill
\resizebox{0.5\linewidth}{!}{
\begin{minipage}{0.625\linewidth}
\begin{algorithm}[H]
    \caption{General Operator Initialization}
    \label{alg:operator_init}
    \begin{algorithmic}[1]
    \For{$j=1,\dots,M$}
        \State {\color{blue} /* Initialize latent batch with encoded $\vy$ \& SDEdit */}
        \State $\epsilon^{(i)}_{T_s} \sim\mathcal{N}(0, I)$ for $i = 1,2,\dots,N$
        \State $\vz_{T_s}^{(i)} = \sqrt{\Bar{\alpha}_{T_s}}\rmE(y) + \sqrt{1-\Bar{\alpha}_{T_s}}\epsilon^{(i)}_{T_s}$
        \State {\color{blue} /* Use MPGD to obtain latent estimations $\{\vz_{0|t}^{(i)}\}_{i=1}^N$ */}
        \For{$t=T_j,\dots,1$ and all $i$ in parallel}
            \State Calculate $\hat{\epsilon}_\theta(\vz_{t}^{(i)},t)$ with Eq.~\ref{eq:cfg}
            \State $\vz_{0|t}^{(i)} = \frac{1}{\sqrt{\Bar{\alpha}_t}}(\vz_{t}^{(i)}-\sqrt{1-\Bar{\alpha}_t}\hat{\epsilon}_\theta(\vz_{t}^{(i)},t))$
            \If{$j \neq 1$}
            \State Update $\vz^{(i)}_{0|t}$ with Eq.~\ref{eq:emprical_mpgd}
            \EndIf
            \State $\vz^{(i)}_{t-1} = \sqrt{\Bar{\alpha}_{t-1}}\vz^{(i)}_{0|t}+\sqrt{1-\Bar{\alpha}_{t-1}-\sigma_{t}^{2}}\epsilon_\theta(\vz_{t}^{(i)}, t)+\sigma_{t}\epsilon_{t}^{(i)}$ for $\epsilon^{(i)}_{t} \sim\mathcal{N}(0, I)$
        \EndFor
        \State {\color{blue} /* Update $\hat{\phi}$ with $\{\vz_{0|t}^{(i)}\}_{i=1}^N$ and Eq.~\ref{eq:phi_loss} */}
        \For{$k=1,\dots,K$}
            \State $\hat{\phi} = \text{Adam}(\hat{\phi}, \frac{1}{N}\sum_{i=1}^NL_\phi(\vz_{0}^{(i)}, \hat{\phi}))$
        \EndFor
    \EndFor
    \State \textbf{return} $\hat{\phi}$
    \end{algorithmic}
\end{algorithm}
\end{minipage}
}
\end{figure}

\subsection{Sampling from the posterior}
\label{sec:posterior}

In addition to a strong prior score, our output image should also satisfy the measurement constraint. Given the problem setup in \Eqref{eq:problem_setting}, the measurements are subject to additive Gaussian noise $\vn$, hence $\log p(\vy|\vx,\gA_\phi) = -\frac{1}{2\sigma^2}\|\vy-\gA_\phi(\vx)\|^2_2$. When $\gA_\phi$ is known, we can use the MPGD update rule in~\Eqref{eq:mpgd} for posterior sampling.

However, since $\gA_\phi$ is unknown, the true parameters $\phi$ is often approximated by another set of parameters $\hat{\phi}$. This approximation is usually addressed by one of the two strategies: an alternating optimization scheme that jointly approximates $\vx$ and $\phi$, or obtaining a reliable $\hat{\phi}$ first then solving a non-blind inverse problem. The first approach is well-suited for training-free settings, but it is often highly sensitive to $\hat{\phi}$ initialization and tuning. The second approach can perform well if a strong $\hat{\phi}$ is obtained, though it usually requires training and additional restrictions.
We propose a hybrid strategy:
we first obtain a reliable initial $\hat{\phi}$, and then perform an alternating optimization to iteratively refine both the operator parameter and data estimations throughout the diffusion process.

\paragraph{$\hat{\phi}$ Initialization}
Sometimes initializing the operator parameters is straightforward -- for example, for a slight blur, an identity function can be a reasonable starting point. Alternatively, reliable operator priors, such as the pre-trained prior in BlindDPS or the Gaussian prior in GibbsDDRM for linear operators, can also provide effective initializations for simple kernels. However, for complex operator estimator like neural networks, obtaining good initializations becomes more challenging.

To address this, we propose a new algorithm for general operator initialization.
Since the goal here is only to initialize $\hat{\phi}$, the quality of intermediate $\vx$ estimates is less critical as long as they provide useful signals. Unlike other diffusion-based methods that alternate optimization targets at each diffusion step, we use SDEdit~\citep{sdedit} and MPGD~\citep{he2024manifold} with very few diffusion steps to quickly obtain a batch of $\vx$ estimates. We then perform maximum likelihood estimation (MLE) on these estimates to update $\hat{\phi}$ using Adam optimizer. As detailed in Algorithm~\ref{alg:operator_init}, repeating this process can leverage the diffusion prior to quickly converge to a reliable starting point for $\hat{\phi}$.

Notice that we only assume that $\phi$ can be approximated by differentiable functions, allowing $\gA_{\hat{\phi}}$ to be parameterized by general model families such as neural networks. Moreover, as mentioned earlier, any well-performing operator priors can be seamlessly integrated into our framework.

\paragraph{Iterative Refinement}
After $\hat{\phi}$ initialization, we perform another alternating optimization to refine both the operator and the image. Throughout the diffusion process, we alternate between updating $\vz_{0|t}$ using MPGD with $\gA_{\hat{\phi}}$ fixed and updating the MLE estimation of $\hat{\phi}$ using Adam with $\vz_{0|t}$ fixed.

Since unlike GibbsDDRM, which uses Langevin dynamics to update the operator, we solve for a local optimum. Therefore, it’s reasonable not to update the operator too frequently. Empirically, we find that periodic updates to the operator combined with time-traveling~\citep{lugmayr2022repaint,freedom} yield the best results.
In addition, since MPGD supports any differentiable loss function, we can incorporate regularization to further improve the visual quality. In practice, we use
\begin{equation}
\small
    \vz_{0|t} = \vz_{0|t}  - c_t\nabla_{\vz_{0|t}} (\|\vy-\gA_{\hat{\phi}} (\rmD(\vz_{0|t}))\|^2_2 + \lambda_\vz \text{LPIPS}(\rmD(\vz_{0|t}), \vy))
    \label{eq:emprical_mpgd}
\end{equation}
as the MPGD update rule and
\begin{equation}
    L_\phi(\vz_{0|t}, \hat{\phi}) = \|\vy-\gA_{\hat{\phi}} (\rmD(\vz_{0|t}))\|^2_2 + \lambda_\phi \|\hat{\phi}\|_1
    \label{eq:phi_loss}
\end{equation}
as the Adam objective. Here $\text{LPIPS}(\cdot)$ denotes the LPIPS distance between two images and $\|\cdot\|_1$ denotes the $L_1$ regularization term. $\lambda_\vz$ and $\lambda_\phi$ are adjustable hyperparameters.

Combining large-scale language-conditioned diffusion priors, effective operator initializations and the iterative refinement process described above, we introduce \modelname{}, a new training-free algorithm for blind inverse problem solving that supports diverse target image distributions and flexible operators.

\section{Experiments}
\label{sec:exps}

\subsection{Experimental Setup}
We empirically verify the performance of our method with two linear deblurring tasks, Gaussian deblurring and motion deblurring, and a non-linear restoration task, JPEG decompression. 

\paragraph{Setup}
We conduct quantitative comparisons on FFHQ 256 $\times$ 256 \citep{karras2019style} and AFHQ-dog 256 $\times$ 256 \citep{Choi_2020_CVPR}. Following~\citet{murata2023gibbsddrm}, we use 1000 images for FFHQ and 500 images for AFHQ.
We use three quantitative metrics: LPIPS \citep{zhang2018unreasonable}, PSNR and KID~\citep{binkowski2018demystifying}.
PSNR directly measures the pixel-by-pixel accuracy of the reconstructed image compared to the original, while LPIPS measures how similar two images are in a way that reflects human perception. Finally, KID evaluates the image fidelity on a distribution level.

\paragraph{Baselines}
We compare our method against other state-of-the-art approaches as baselines. Specifically, we choose Pan-l0 \citep{panl0} and Pan-DCP \citep{pandcp} as the optimization-based method, SelfDeblur \citep{selfdeblur} as the self-supervised approach, PRNet \citep{mprnet} and DeblurGANv2 \citep{deblurganv2} as the supervised baselines, and BlindDPS \citep{blinddps}, BIRD~\citep{chihaoui2024blind} and GibbsDDRM \citep{murata2023gibbsddrm} as the diffusion-based methods.
All baselines are experiments using their open-sourced code and pre-trained models.

\paragraph{Implementation Details}
\label{par: implementation}
We conduct all experiments on an NVIDIA A6000 GPU.
We use Stable Diffusion 1.4~\citep{rombach2021highresolution}, DDIM $200$-step, and $T_s=150, M=4$ for all tasks. We set positive prompts as \textit{``a clear headshot of a person"} and \textit{``a clear headshot of an animal"} for FFHQ and AFHQ experiments respectively. We adjust negative prompts according to each individual task and the details are be provided in the Appendix~\ref{app:implementation}.

\begin{table}[t]
\small
\centering
\caption{Quantitative results on blind linear deblurring tasks. ``GibbsDDRM'' denotes the results from directly running their open-sourced code and ``GibbsDDRM*'' denotes the results after adjusting their code to match with the ground truth kernel padding.}
\label{tab:linear_results}
\resizebox{\textwidth}{!}{
\begin{tabular}{@{}ccccccccccccc@{}}
\toprule
\multicolumn{1}{l}{}    & \multicolumn{6}{c}{FFHQ}                                                                             & \multicolumn{6}{c}{AFHQ}                                                                              \\ \midrule
\multirow{2}{*}{Method} & \multicolumn{3}{c}{Motion}                       & \multicolumn{3}{c}{Gaussian}                      & \multicolumn{3}{c}{Motion}                        & \multicolumn{3}{c}{Gaussian}                      \\ \cmidrule(l){2-13} 
                        & LPIPS $\downarrow$         & PSNR $\uparrow$           & KID $\downarrow$             & LPIPS $\downarrow$          & PSNR $\uparrow$           & KID $\downarrow$             & LPIPS $\downarrow$          & PSNR $\uparrow$           & KID $\downarrow$             & LPIPS $\downarrow$          & PSNR$\uparrow$& KID $\downarrow$             \\ \midrule
DPS w/ GT kernel                & 0.164         & 22.82          & 0.0046          & 0.138          & 24.48          & 0.0052          & 0.367          & 21.28          & 0.1120          & 0.330           & 23.52          & 0.0813          \\ 
\midrule
SelfDeblur              & 0.732         & 9.05           & 0.1088          & 0.733          & 8.87           & 0.0890           & 0.742          & 9.04           & 0.0650           & 0.736          & 8.84           & 0.0352          \\
MPRNet                  & 0.292         & \underline{22.42}        & 0.0467          & 0.334          & 23.23          & 0.0511          & 0.324          & \underline{22.09}          & 0.0382          & 0.379          & 21.97          & 0.0461          \\
DeblurGANv2               & 0.309         & \textbf{22.55} & 0.0411          & 0.325          & 26.61 & 0.0227          & 0.323          & \textbf{22.74} & 0.0350           & 0.340           & \textbf{27.12} & 0.0073          \\
Pan-l0                  & 0.389         & 14.10           & 0.1961          & 0.265          & 20.68          & 0.1012          & 0.414          & 14.16          & 0.1590           & 0.276          & 21.04          & 0.0320           \\
Pan-DCP                 & 0.325         & 17.64          & 0.1323          & 0.235          & 24.93          & 0.0490           & 0.371          & 17.63          & 0.1377          & 0.297          & \underline{25.11}          & 0.0263          \\
BlindDPS                & 0.246         & 20.93          & 0.0153          & 0.216          & 25.96          & \underline{0.0205}       & 0.393          & 20.14          & 0.0913          & 0.330           & 24.79          & 0.0268          \\ 
BIRD                & 0.294         & 19.23          & 0.0491          & 0.212          & 21.95          & 0.0414       & 0.438          & 18.92          & 0.0286          & 0.320           & 21.87          & 0.0089          \\ 
GibbsDDRM                & 0.293         & 20.52          & 0.0746          & 0.216          & \underline{27.03}          & 0.0430          & 0.303          & 19.44          & 0.0265          & 0.257           & 24.01          & \textbf{0.0040}          \\ \midrule
\textbf{\modelname{} (Ours)}           & \underline{0.230} & 20.96          & \textbf{0.0084} & \underline{0.197} & 21.08          & \textbf{0.0068} & \textbf{0.262} & 21.20           & \textbf{0.0132} & \textbf{0.204} & 24.33          & \underline{0.0065} \\ \midrule
GibbsDDRM*                & \textbf{0.199}         & 22.36          & \underline{0.0309}         & \textbf{0.155}          & \textbf{27.65}          & 0.0252          & \underline{0.278}          & 19.00          & \underline{0.0180}          & \underline{0.224}           & 21.62          & \textbf{0.0040}          \\ 
\bottomrule
\end{tabular}
}
\end{table}

\begin{figure}[t]
    \centering
    \includegraphics[width=\textwidth]{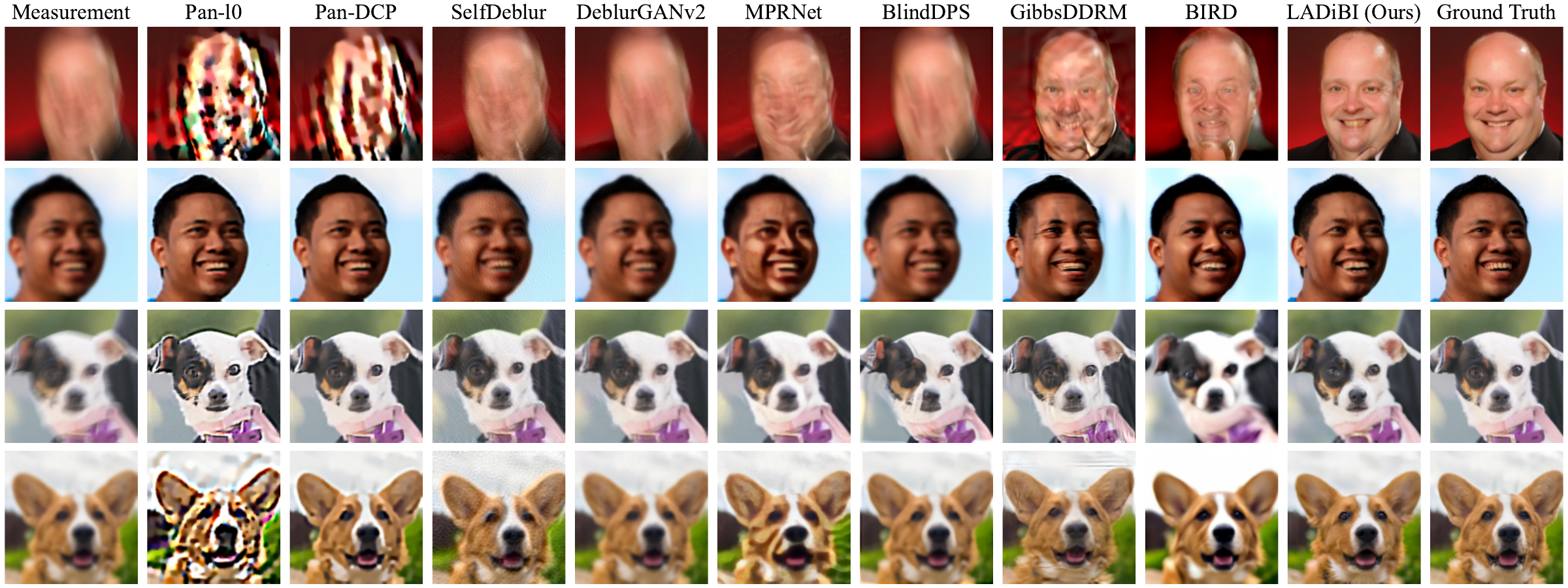}
    \caption{Qualitative results on blind linear deblurring tasks. From top to bottom we showcase examples from motion deblur on FFHQ, Gaussian deblur on FFHQ, motion deblur on AFHQ, and Gaussian deblur on AFHQ respectively.}
    \label{fig:linear_fig}
\end{figure}

\subsection{Blind Linear Deblurring}
We first conduct experiments on linear deblurring, which is the design space of most baselines. We evaluate all methods on two blurring kernels: Gaussian blur and motion blur.
We apply random motion blur kernels with intensity $0.5$ and Gaussian blur kernels with standard deviation $3$. Measurements are derived by applying a pixel-wise Gaussian noise with $\sigma=0.02$.
We use a $61\times61$ convolutional matrix as $\hat{\phi}$ and initialize it as a Gaussian kernel of intensity $6.0$ following~\citet{murata2023gibbsddrm}

As shown in Table~\ref{tab:linear_results} and Figure~\ref{fig:linear_fig}, our method matches the performance of the state-of-the-art method GibbsDDRM, which is explicitly designed to solve linear problems using SVDs. In fact, GibbsDDRM’s highly specialized design makes it so sensitive that even small discrepancies (e.g. kernel padding) between their modeling assumption and the ground truth operator can result in significant performance degradation. Moreover, although BlindDPS and GibbsDDRM use diffusion models that are trained on the exact data distributions tested, our method can still match or outperform them with the large-scale pre-trained model. While supervised methods obtain higher PSNR, our method produces the fewest artifacts, validated by LPIPS and KID scores. Overall, our method offers a competitive performance on linear tasks,
even though it is designed for more general applications.

\subsection{Blind JPEG Decompression}
We further compare all methods on JPEG decompression, a particularly challenging task due to its non-linear, non-differentiable nature. Unlike traditional settings, our experiments provide no task-specific knowledge, such as the compression algorithm, quantization table, or factors -- algorithms rely solely on the measurement images.

\label{par:jpeg-setup}
We generate measurements using JPEG compression with a quantization factor of 2. 
We use a neural network with a 3-layer U-net~\citep{ronneberger2015unetconvolutionalnetworksbiomedical} to parametrize the operator.
The operator is initialized using Algorithm \ref{alg:operator_init} with $M=8$ and $N=4$.

\begin{table}[t]
\small
\centering
\caption{Quantitative results (the mean and standard deviation) on blind JPEG decompression task.}
\label{tab:jpeg_results}
\resizebox{\textwidth}{!}{
\begin{tabular}{@{}ccccccc@{}}
\toprule
\multicolumn{1}{c}{\multirow{2}{*}{Method}} & \multicolumn{3}{c}{FFHQ}                          & \multicolumn{3}{c}{AFHQ}                          \\ \cline{2-7} 
\multicolumn{1}{c}{}                        & LPIPS $\downarrow$          & PSNR $\uparrow$          & KID $\downarrow$            & LPIPS  $\downarrow$        & PSNR  $\uparrow$         & KID  $\downarrow$           \\ \hline
Pan-I0      & 0.787 $\pm$ 0.084 & 12.72 $\pm$ 2.53 & 0.2189 & 0.825 $\pm$ 0.083 & 13.15 $\pm$ 2.68 & 0.2168 \\
Pan-DCP     & 0.710 $\pm$ 0.067 & 14.91 $\pm$ 2.63 & 0.2318 & 0.673 $\pm$ 0.055 & 18.27 $\pm$ 1.60 & 0.2472 \\
SelfDeblur  & 0.676 $\pm$ 0.054 &  8.84 $\pm$ 1.93 & 0.1959 & 0.703 $\pm$ 0.052 &  8.89 $\pm$ 1.72 & 0.1173 \\
MPRNet      & 0.785 $\pm$ 0.048 &  5.99 $\pm$ 1.81 & 0.8008 & 0.769 $\pm$ 0.048 &  5.93 $\pm$ 1.78 & 0.6809 \\
DeblurGAN   & 0.473 $\pm$ 0.058 & \underline{21.48} $\pm$ 1.85 & 0.2207 & 0.502 $\pm$ 0.061 & \textbf{21.76} $\pm$ 1.84 & 0.1527 \\
BlindDPS    & 0.431 $\pm$ 0.079 & \textbf{21.55} $\pm$ 1.96 & 0.1791 & \underline{0.397} $\pm$ 0.069 & 20.87 $\pm$ 1.89 & 0.2108 \\
BIRD        & \underline{0.406} $\pm$ 0.047 & 20.68 $\pm$ 1.09 & \underline{0.0525} & 0.425 $\pm$ 0.068 & 21.08 $\pm$ 1.50 & \underline{0.0673} \\
GibbsDDRM   & 0.841 $\pm$ 0.057 & 13.91 $\pm$ 1.26 & 0.2915 & 0.775 $\pm$ 0.059 & 14.59 $\pm$ 1.48 & 0.2915 \\ \hline
Ours        & \textbf{0.268} $\pm$ 0.070 & 21.40 $\pm$ 1.18 & \textbf{0.0172} & \textbf{0.315} $\pm$ 0.075 & \underline{21.12} $\pm$ 1.27 & \textbf{0.0216} \\ \bottomrule
\end{tabular}
}
\end{table}

\begin{figure}[t]
    \centering
    \includegraphics[width=\textwidth]{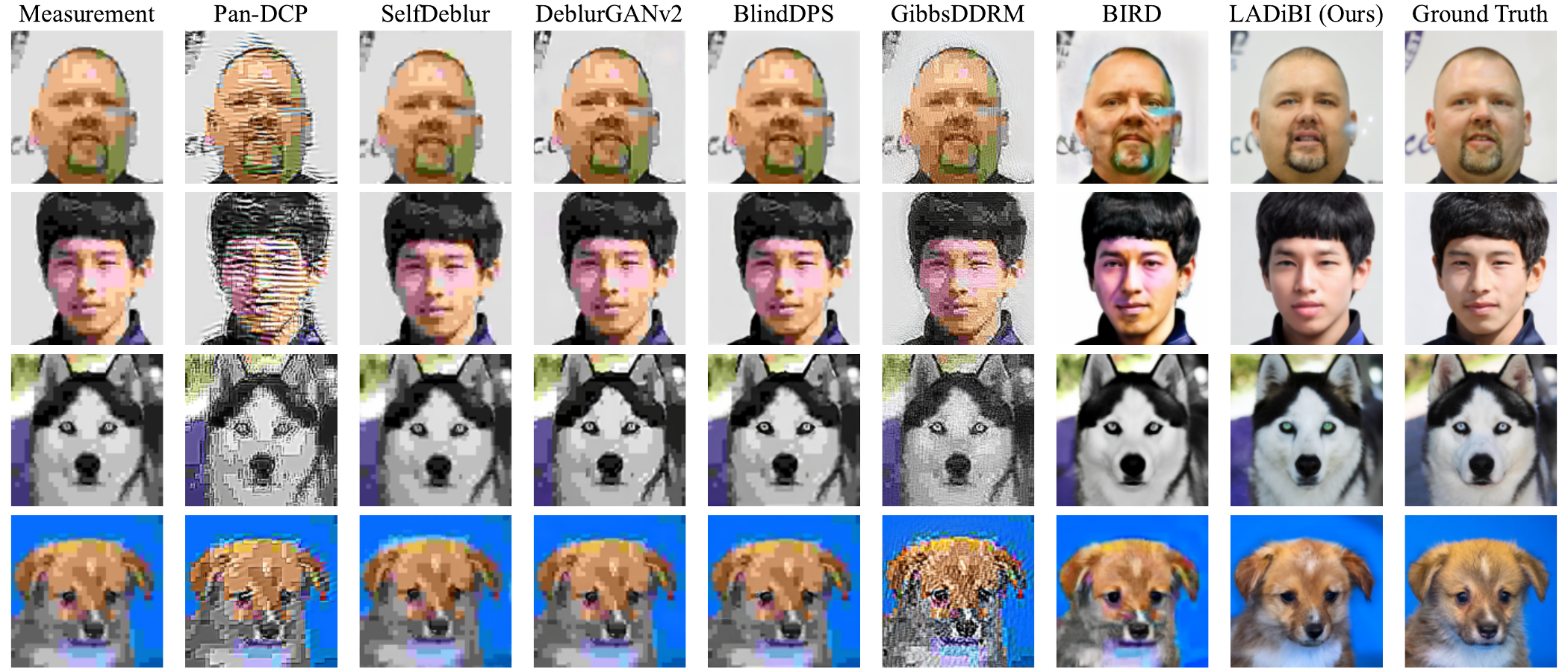}
    \caption{Qualitative results on the blind JPEG decompression task.}
    \label{fig:jpeg_fig}
\end{figure}

Table \ref{tab:jpeg_results} and Figure~\ref{fig:jpeg_fig} present the results on the JPEG decompression task. It is evident through both quantitative and qualitative results that our method is the only one capable of enhancing the fidelity of the images and maintaining consistency to measurements.
Unlike the baselines, which struggle with this task due to their limited posterior formulations, our flexible framework adapts to approximate this challenging operator and produces high quality images.

\subsection{Ablation Study}
\begin{figure}[t]
  \centering
  \begin{minipage}{0.5\linewidth}
    \small
\centering
\captionof{table}{Ablation study on JPEG decompression.}
\label{tab:ablation}
\resizebox{\columnwidth}{!}{
\begin{tabular}{@{}cccc@{}}
\toprule
Ablation & LPIPS $\downarrow$          & PSNR $\uparrow$          & KID $\downarrow$    \\ \hline
    W/o text prompts               & 0.508 & 19.30  & 0.0319 \\
    W/o negative prompts           & 0.440  & 19.44 & 0.0242 \\
    Using generic $\vc_-$      & 0.425 & 19.48 & 0.0255 \\ 
    Using task-irrelevant $\vc_-$    & 0.406 & 19.40  & 0.0237 \\
    W/o MPGD guidance                  & 0.403 & 19.44 & 0.0274 \\ 
    W/o regularization              & 0.307 & \textbf{21.35} & 0.0172 \\ 
    W/o $\hat{\phi}$ Initialization            & \underline{0.289} & 20.17 & \underline{0.0170} \\ \hline
    \textbf{\modelname{}}                    & \textbf{0.262} & \underline{21.20}  & \textbf{0.0132} \\
    \bottomrule
\end{tabular}
}
  \end{minipage}
  \hfill %
  \begin{minipage}{0.475\linewidth}
    \centering
    \includegraphics[width=0.9\columnwidth]{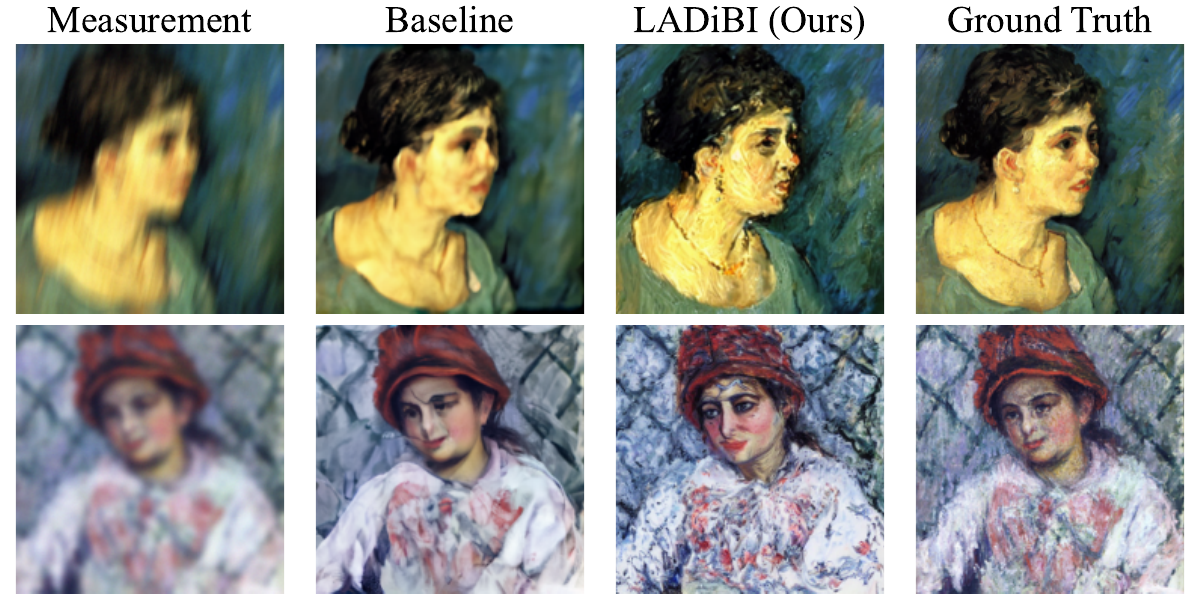}
    \caption{Qualitative results on blind deblurring Monet paintings.}
    \label{fig: monet_fig}
  \end{minipage}
\vspace{-10pt}
\end{figure}

To showcase the effectiveness of each part of our algorithm, we conduct ablation study on AFHQ dataset with JPEG decompression task. In particular, we test the importance of suitable ad-hoc prompts, the use of MPGD guidance, the MPGD regularization term, and the operator initialization for neural network $\gA_{\hat{\phi}}$. We keep SDEdit in all settings to encode the measurement information for fair comparisons. As shown in Table \ref{tab:ablation}, each of the aforementioned components plays a significant role in our scheme and is indispensable for producing high quality results.

\subsection{Style Preserving Deblurring}

In order to show our method's applicability on a wider range of image and operator distributions, we present indicative examples of Gaussian and motion deblurring on Monet paintings in Figure \ref{fig: monet_fig}. Additional operators and image distributions demonstrations are available in Appendix~\ref{app:demo}.
We use the same negative prompts as in the previous experiments and \textit{``a portrait of a person as a Monet style painting"} as positive prompts. We notice that, although the images portray human faces, the baseline method using models trained on realistic human face images cannot accurately solve the problem, while our method effectively generates images consistent with both the measurement image and the painting style present in the ground truth image.

\section{Conclusion}
\label{sec:conclusion}

In this work, we propose \modelname{}, a new training-free algorithm to solving blind inverse problems in image restoration using large-scale pre-trained text-to-image diffusion models. With unknown degradation operators, our method leverages text prompts as well as posterior guidance on intermediate diffusion steps to restore desired images based on the measurements. Experiments demonstrate that \modelname{}'s effectiveness on diverse operator and image distributions.

\section*{Acknowledgment}
This work is sponsored by Sony AI and is also supported in part by the ONR grant N000142312368.

\bibliography{example_paper}
\bibliographystyle{abbrvnat}

\newpage
\appendix
\onecolumn
\section{Additional Implementation Details}
\label{app:implementation}
In this section, we offer more detailed information regarding the implementation setup for Algorithms 1 and 2. All experiments are conducted using an NVIDIA A6000 GPU. In an effort to make our methodology generalizable and extensible to other inverse problems, we attempt to maintain the same hyper-parameter values across tasks wherever possible. Each hyper-parameter value has been chosen after conducting preliminary experiments for a specific range and opting for the value that offers the best performance. Indicative examples of such experiments are shown in \ref{sec:abl_appdx}.

In particular, we run Algorithm 1 for $T_s=150$ timesteps while performing 4 repetitions as part of the time-traveling strategy. We encode the measurement as $\vx_{T_s}$ by applying the forward diffusion process up to timestep 800. In parallel with the reverse diffusion process, we update $\hat{\phi}$ every 5 timesteps, each time conducting $K=150$ gradient steps. We adjust the $c_t$ and the $\lambda_{\phi}$ parameters of Equations 8 and 9 to 30 and 2 respectively. In terms of the operator initialization process described by Algorithm 2, we make use of a batch of $N=4$ samples and run $M=8$ iterations, each comprising $T_j=60$ timesteps.

We also take into consideration that the targets and measurements reside in the $256 \times 256$ pixel space, whereas Stable Diffusion v1 operates on images of pixel size $512 \times 512$. To address this disparity, we initially upsample the measurement using bilinear interpolation in order to transform it to a $512 \times 512$ image, and then downsample the resulting $x_0$ to map the final estimate back to the original image space.

\begin{figure*}[t]
    \centering
    \includegraphics[width=0.8\textwidth]{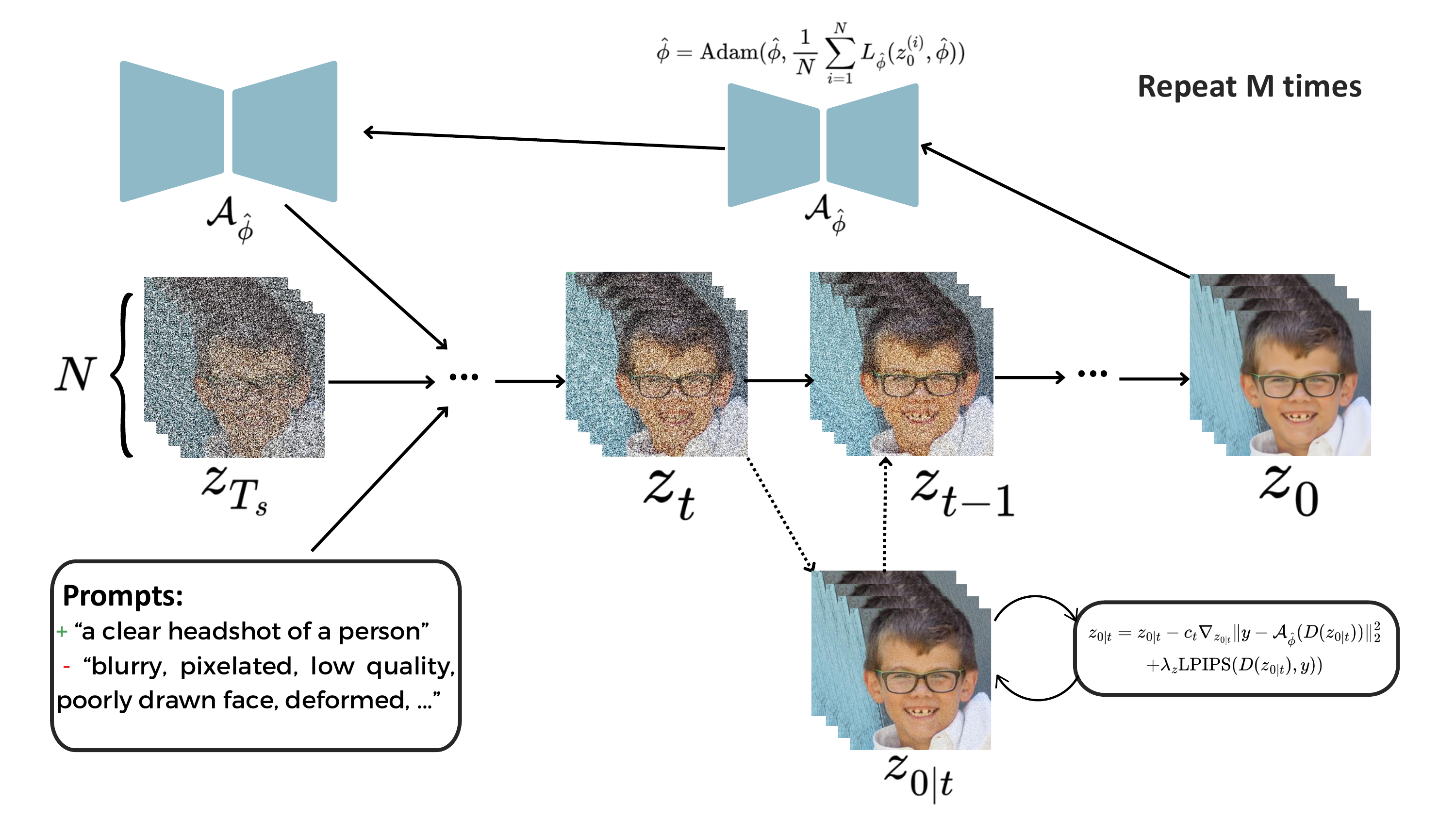}
    \caption{A schematic overview of general operator initialization (Algorithm~\ref{alg:operator_init}).}
    \label{fig:alg2_fig}
\end{figure*}

The schematic overviews of Algorithms 1 and 2 are presented in Figures \ref{fig:alg1_fig} and \ref{fig:alg2_fig}.

In addition, there are some parameters in our approach for which employing a task-aware setup strategy is essential for state-of-the-art performance. These are the prompt sentences as well as the architecture of $\hat{{\phi}}$. Here we provide details with respect to these parameters according to each restoration task:

\paragraph{Motion Deblurring} 
\begin{itemize}
    \item Positive prompts: \textit{``a clear headshot of a person/animal"}
    \item Negative prompts: \textit{``shaken image, motion blur, pixelated, lowres, text, error, cropped, worst quality, blurry ears, low quality, ugly, duplicate, morbid, mutilated, poorly drawn face, mutation, deformed, blurry, dehydrated, blurry hair, bad anatomy, bad proportions, disfigured, gross proportions"}
    \item $\hat{\phi}$ architecture: A single $61\times 61$ convolutional block with 3 input and 3 output channels.
\end{itemize}

\paragraph{Gaussian Deblurring} 
\begin{itemize}
    \item Positive prompts: \textit{``a clear headshot of a person/animal"}
    \item Negative prompts: \textit{``blurry, gaussian blur, lowres, text, error, cropped, worst quality, blurry ears, low quality, ugly, duplicate, morbid, mutilated, text in image, DSLR effect, poorly drawn face, mutation, deformed, dehydrated, blurry hair,
   bad anatomy, bad proportions, disfigured, gross proportions"}
    \item $\hat{\phi}$ architecture: A single $61\times 61$ convolutional block with 3 input and 3 output channels.
\end{itemize}

\paragraph{JPEG Decompression} 
\begin{itemize}
    \item Positive prompts: \textit{``a clear headshot of a person/animal"}
    \item Negative prompts: \textit{``pixelated, lowres, text, error, cropped, worst quality, blurry ears, low quality, jpeg artifacts, ugly, duplicate, morbid, mutilated, text in image, DSLR effect, poorly drawn face, mutation, deformed, blurry, dehydrated, blurry hair,
   bad anatomy, bad proportions, disfigured, gross proportions"}
    \item $\hat{\phi}$ architecture: A neural network with a typical 3-layer U-net~\citep{ronneberger2015unetconvolutionalnetworksbiomedical} architecture. Each layer consists of 2 convolutional blocks of size $3\times3$ with ReLU activations and number of input and output channels ranging from 32 to 128.
\end{itemize}

We conduct quantitative experiments on FFHQ $256\times 256$~\citep{karras2019style} and AFHQ $256\times 256$~\citep{Choi_2020_CVPR}. Images in FFHQ are publicized under Creative Commons BY 2.0, Creative Commons BY-NC 2.0, Public Domain Mark 1.0, Public Domain CC0 1.0, or U.S. Government Works license and AFHQ is publicized under Attribution-NonCommercial 4.0 International license.

% The code implementation for \modelname{} is linked \href{https://anonymous.4open.science/r/LADiBI-NeurIPS2025/}{\color{magenta}here}.

\section{Additional Results}

\begin{figure}[t]
    \centering
    \includegraphics[width=0.95\textwidth]{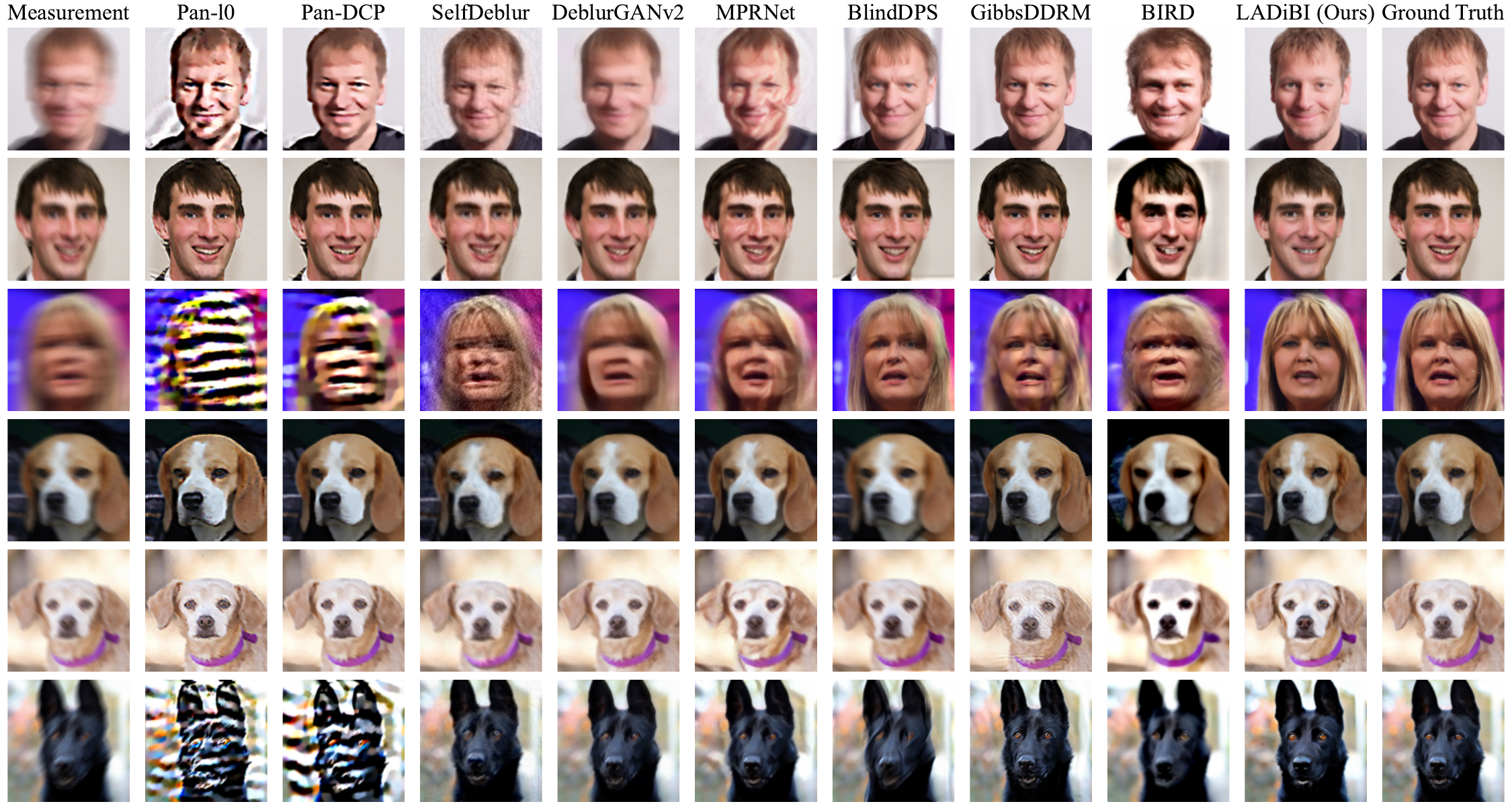}
    \caption{Additional qualitative results on motion deblurring task.}
    \label{fig:motion_fig_appdx}
\end{figure}

\begin{figure}[h]
    \centering
    \includegraphics[width=0.95\textwidth]{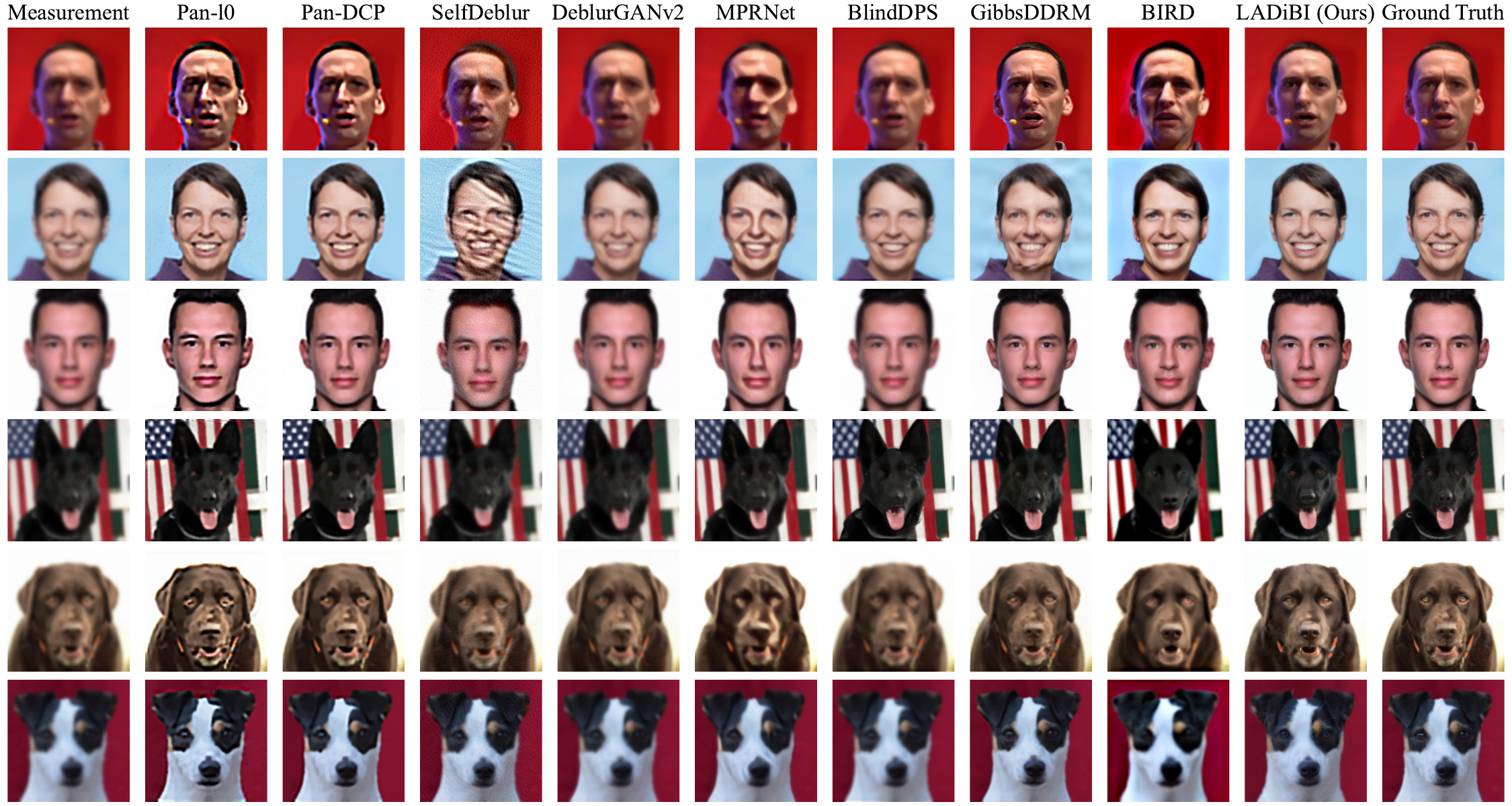}
    \caption{Additional qualitative results on Gaussian deblurring task.}
    \label{fig:gaussian_fig_appdx}
\end{figure}

\begin{figure}[t]
    \centering
    \includegraphics[width=0.95\textwidth]{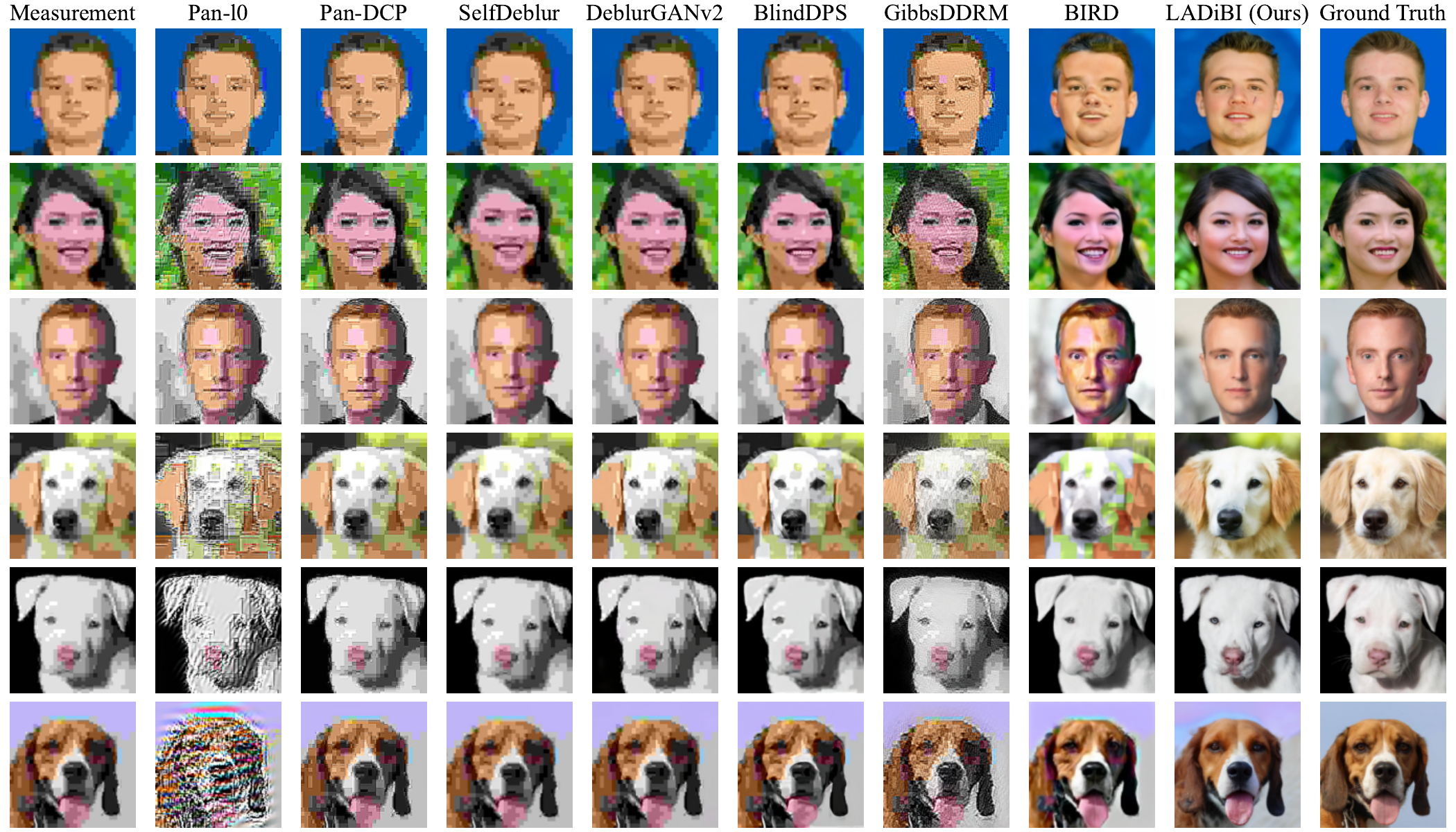}
    \caption{Additional qualitative results on JPEG decompression task.}
    \label{fig:jpeg_fig_appdx}
\end{figure}

\begin{figure}[t]
    \centering
    \includegraphics[width=0.95\textwidth]{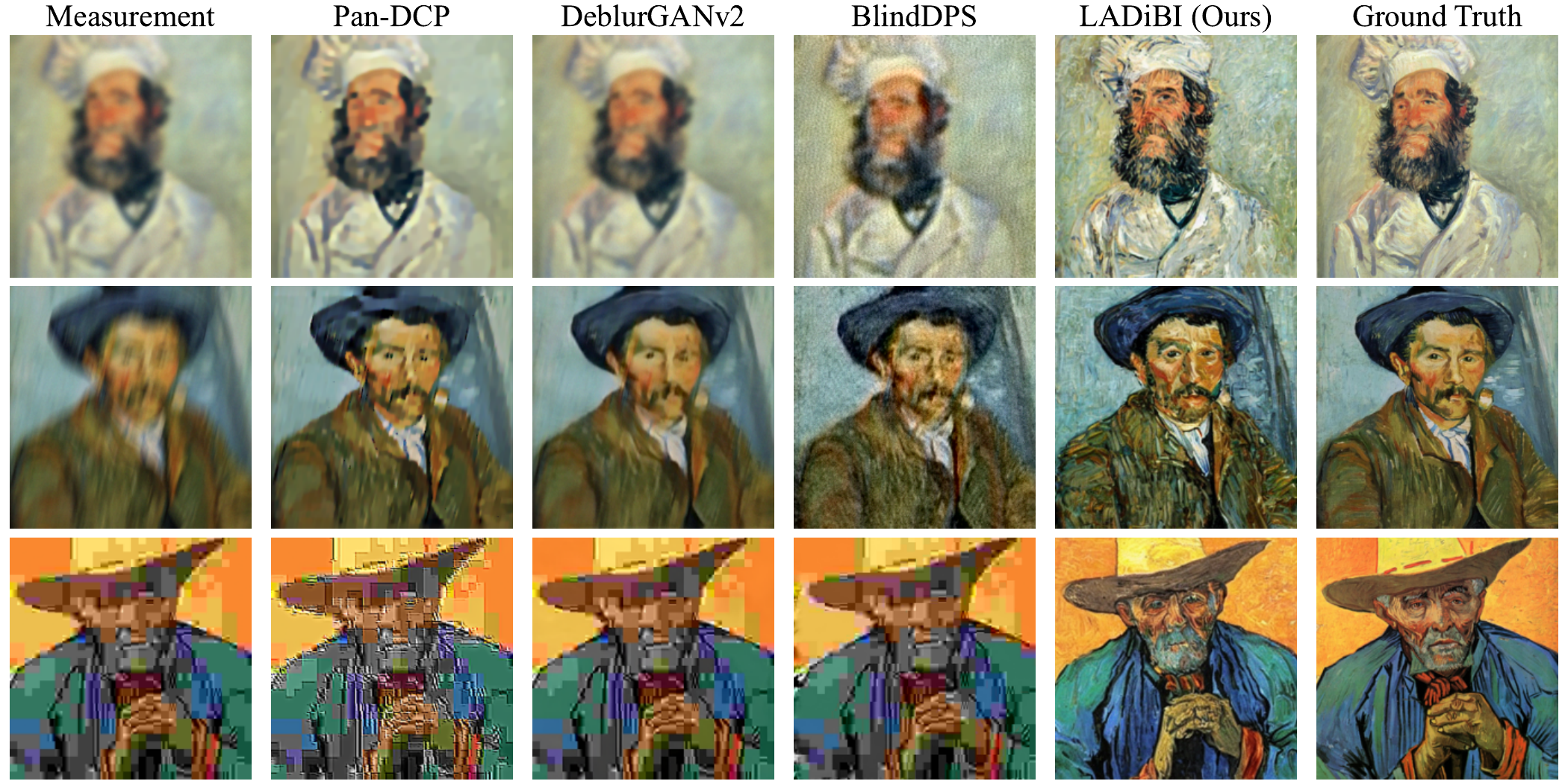}
    \caption{Additional qualitative results on painting restoration.}
    \label{fig:painting_fig_appdx}
\end{figure}

In this section, we present additional experimental results on all benchmarks as well as additional tasks and data distributions. Table~\ref{tab:linear_ffhq_full} include the standard deviation of the linear benchmark results.
Figures \ref{fig:motion_fig_appdx}, \ref{fig:gaussian_fig_appdx} and \ref{fig:jpeg_fig_appdx} offer more qualitative comparisons against baseline methods on motion deblurring, gaussian deblurring and JPEG decompression respectively. In Figure \ref{fig:painting_fig_appdx}, we include comparisons against selected baseline methods on the 3 benchmark tasks using a set of Monet and Van Gogh portrait paintings as ground truth images.

\begin{figure}[t]
    \centering
    \begin{minipage}{0.48\textwidth}
        \centering
        \includegraphics[width=\columnwidth]{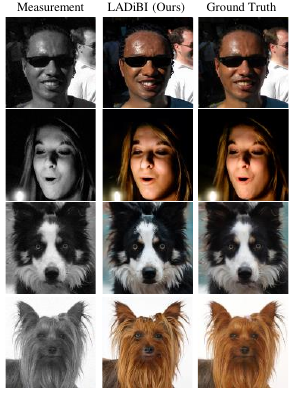}
    \caption{Qualitative results on colorization task.}
    \label{fig: colorization_fig}
     \end{minipage}
    \hfill
    \begin{minipage}{0.48\textwidth}
        \centering
        \includegraphics[width=\columnwidth]{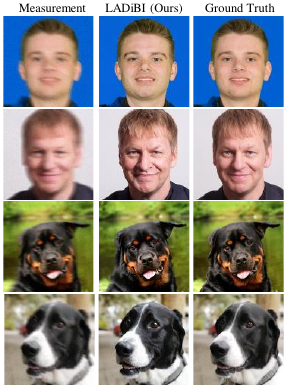}
    \caption{Qualitative results on non-linear deblurring task.}
    \label{fig: nonlineardeblur_fig}
      \end{minipage}
\end{figure}

\begin{figure}[t]
    \centering
    \begin{minipage}{0.48\textwidth}
        \centering
    \includegraphics[width=\columnwidth, height=15cm]{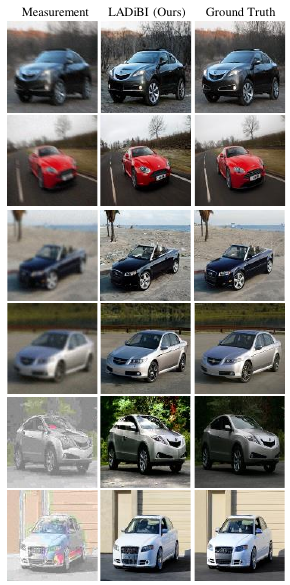}
    \caption{Demo inverse problem solving on car images.}
    \label{fig: cars_fig}
     \end{minipage}
    \hfill
    \begin{minipage}{0.48\textwidth}
        \centering
    \includegraphics[width=\columnwidth, height=15cm]{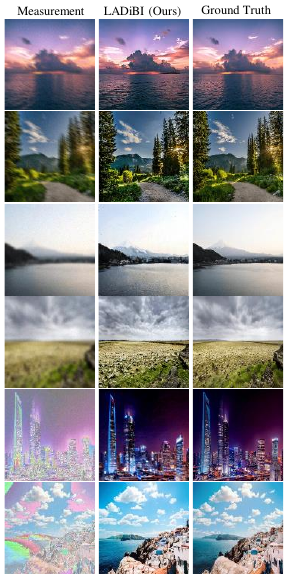}
    \caption{Demo inverse problem solving on landscape images.}
    \label{fig: landscapes_fig}
      \end{minipage}
\end{figure}

\label{app:demo}
We also show additional demonstrations of applying \modelname{} to solve colorization and non-linear deblurring problems. Figures \ref{fig: colorization_fig} and \ref{fig: nonlineardeblur_fig} substantiate our claim that our algorithmic scheme is flexible enough to adapt to various inverse tasks.
Additionally, thanks to the broad range of image distributions encapsulated by the pretrained model, as well as the minimal assumptions imposed by the proposed methodology, our approach is able to accurately solve inverse problems for a broad range of images. Figures \ref{fig: cars_fig} and \ref{fig: landscapes_fig} demonstrate our method's capability to perform Gaussian deblurring, motion deblurring as well as JPEG decompression for images depicting cars and landscapes.

\begin{figure}[t]
    \centering
    \includegraphics[width=0.7\textwidth]{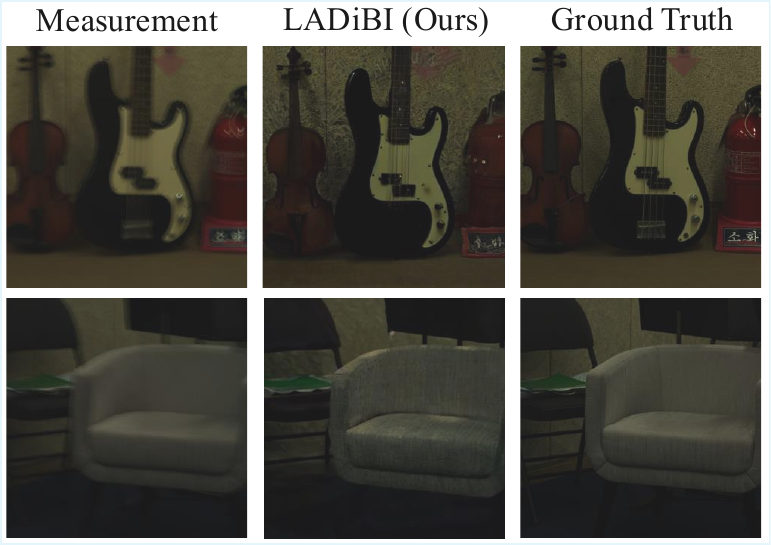}
    \caption{Qualitative samples on the RealBlur dataset.}
    \label{fig:realblur_fig}
\end{figure}

Finally, to assess the real-world performance of our method, we test it on samples from the RealBlur dataset \citep{rim_2020_ECCV}, which contains naturally captured blurry images using GoPro in real life settings. This dataset presents realistic blur patterns that are significantly more complex and diverse compared to the synthetically generated benchmarks.
Figure \ref{fig:realblur_fig} presents qualitative demos, which highlight our method's ability to recover sharp structures and textures under various real-world blur conditions. These results demonstrate the applicability of our approach to real inverse problem solving tasks and support its generalization capability.

\begin{table}[t]
\centering
\small
\caption{Quantitative results on motion and Gaussian deblurring on FFHQ dataset. Mean $\pm$ standard deviation shown.}
\label{tab:linear_ffhq_full}
\begin{tabular}{lcccccc}
\toprule
\multirow{2}{*}{Method} 
  & \multicolumn{3}{c}{Motion} 
  & \multicolumn{3}{c}{Gaussian} \\
\cmidrule(lr){2-4} \cmidrule(lr){5-7}
  & LPIPS $\downarrow$ 
  & PSNR $\uparrow$ 
  & KID $\downarrow$ 
  & LPIPS $\downarrow$ 
  & PSNR $\uparrow$ 
  & KID $\downarrow$ \\
\midrule
DPS w/ GT kernel 
  & 0.164 
  & 22.82 
  & 0.0046 
  & 0.138 
  & 24.48 
  & 0.0052 \\
\midrule

SelfDeblur 
  & $0.732\pm0.147$ 
  & $ 9.05\pm2.46$ 
  & 0.1088 
  & $0.733\pm0.094$ 
  & $ 8.87\pm2.21$ 
  & 0.0890 \\

MPRNet 
  & $0.292\pm0.101$ 
  & $\underline{22.42}\pm3.21$ 
  & 0.0467 
  & $0.334\pm0.068$ 
  & $23.23\pm2.19$ 
  & 0.0511 \\

DeblurGANv2 
  & $0.309\pm0.111$ 
  & $\mathbf{22.55}\pm3.44$ 
  & 0.0411 
  & $0.325\pm0.145$ 
  & $26.61\pm3.28$ 
  & 0.0227 \\

Pan-l0 
  & $0.389\pm0.090$ 
  & $14.10\pm2.51$ 
  & 0.1961 
  & $0.265\pm0.082$ 
  & $20.68\pm3.81$ 
  & 0.1012 \\

Pan-DCP 
  & $0.325\pm0.105$ 
  & $17.64\pm3.63$ 
  & 0.1323 
  & $0.235\pm0.066$ 
  & $24.93\pm3.60$ 
  & 0.0490 \\

BlindDPS 
  & $0.246\pm0.077$ 
  & $20.93\pm2.09$ 
  & 0.0153 
  & $0.216\pm0.076$ 
  & $25.96\pm2.45$ 
  & 0.0205 \\

BIRD 
  & $0.294\pm0.076$ 
  & $19.23\pm1.88$ 
  & 0.0491 
  & $0.212\pm0.055$ 
  & $21.95\pm1.62$ 
  & 0.0414 \\

GibbsDDRM 
  & $0.293\pm0.099$ 
  & $20.52\pm2.81$ 
  & 0.0746 
  & $0.216\pm0.046$ 
  & $\underline{27.03}\pm1.87$ 
  & 0.0430 \\

\midrule
LADiBI (Ours) 
  & $\underline{0.230}\pm0.076$ 
  & $20.96\pm2.34$ 
  & $\mathbf{0.0084}$ 
  & $\underline{0.197}\pm0.071$ 
  & $21.08\pm2.71$ 
  & $\mathbf{0.0068}$ \\
\midrule

GibbsDDRM* 
  & $\textbf{0.199}\pm0.110$ 
  & $22.36\pm3.79$ 
  & \underline{0.0309}
  & $\textbf{0.155}\pm0.049$ 
  & $\textbf{27.65}\pm2.66$ 
  & \underline{0.0252} \\
\bottomrule
\end{tabular}
\end{table}

\begin{table}[t]
\centering
\small
\caption{Quantitative results on motion and Gaussian deblurring on AFHQ. Mean $\pm$ standard deviation shown.}
\label{tab:quant2}
\begin{tabular}{lcccccc}
\toprule
\multirow{2}{*}{Method} 
  & \multicolumn{3}{c}{Motion} 
  & \multicolumn{3}{c}{Gaussian} \\
\cmidrule(lr){2-4} \cmidrule(lr){5-7}
  & LPIPS $\downarrow$ 
  & PSNR $\uparrow$ 
  & KID $\downarrow$ 
  & LPIPS $\downarrow$ 
  & PSNR $\uparrow$ 
  & KID $\downarrow$ \\
\midrule
DPS w/ GT kernel 
  & 0.367 
  & 21.28 
  & 0.1120 
  & 0.330 
  & 23.52 
  & 0.0813 \\\midrule

SelfDeblur 
  & $0.742\pm0.158$ 
  & $ 9.04\pm1.84$ 
  & 0.0650 
  & $0.736\pm0.112$ 
  & $ 8.84\pm1.45$ 
  & 0.0352 \\

MPRNet 
  & $0.324\pm0.095$ 
  & $\underline{22.09}\pm3.01$ 
  & 0.0382 
  & $0.379\pm0.081$ 
  & $21.97\pm2.55$ 
  & 0.0461 \\

DeblurGANv2 
  & $0.323\pm0.105$ 
  & $\mathbf{22.74}\pm2.89$ 
  & 0.0350 
  & $0.340\pm0.084$ 
  & $\mathbf{27.12}\pm2.94$ 
  & 0.0073 \\

Pan-l0 
  & $0.414\pm0.133$ 
  & $14.16\pm3.97$ 
  & 0.1590 
  & $0.276\pm0.079$ 
  & $21.04\pm3.39$ 
  & 0.0320 \\

Pan-DCP 
  & $0.371\pm0.147$ 
  & $17.63\pm5.94$ 
  & 0.1377 
  & $0.297\pm0.086$ 
  & $\underline{25.11}\pm3.68$ 
  & 0.0263 \\

BlindDPS 
  & $0.393\pm0.061$ 
  & $20.14\pm1.67$ 
  & 0.0913 
  & $0.330\pm0.057$ 
  & $24.79\pm1.76$ 
  & 0.0268 \\

BIRD 
  & $0.438\pm0.110$ 
  & $18.92\pm2.04$ 
  & 0.0286 
  & $0.320\pm0.079$ 
  & $21.87\pm1.95$ 
  & 0.0089 \\

GibbsDDRM 
  & $0.303\pm0.114$ 
  & $19.44\pm3.58$ 
  & 0.0265 
  & $0.257\pm0.121$ 
  & $24.01\pm4.77$ 
  & $\mathbf{0.0040}$ \\

\midrule
LADiBI (Ours) 
  & $\mathbf{0.262}\pm0.096$ 
  & $21.20\pm2.72$ 
  & $\mathbf{0.0132}$ 
  & $\mathbf{0.204}\pm0.066$ 
  & $24.33\pm1.73$ 
  & \underline{0.0065} \\\midrule

GibbsDDRM* 
  & $\underline{0.278}\pm0.099$ 
  & $19.00\pm3.09$ 
  & \underline{0.0180} 
  & $\underline{0.224}\pm0.091$ 
  & $21.62\pm3.53$ 
  & \textbf{0.0040} \\
\bottomrule
\end{tabular}
\end{table}

\clearpage
\section{Additional Ablation Study}
\label{sec:abl_appdx}

\begin{figure}[t]
    \centering
    \begin{minipage}{0.48\textwidth}
        \centering
        \includegraphics[width=\textwidth]{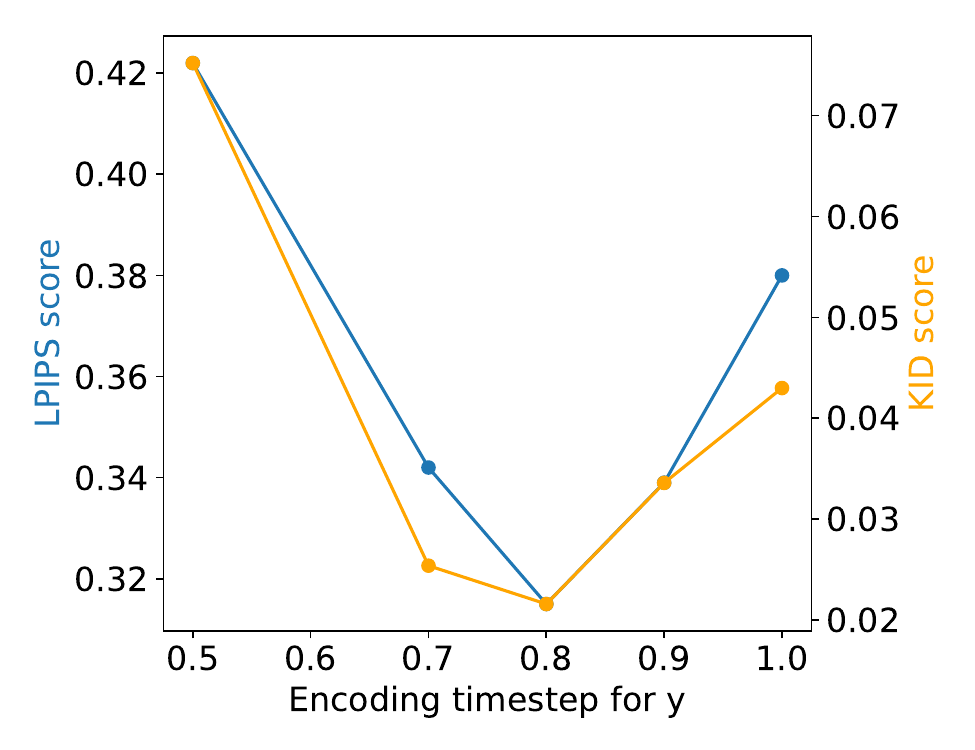}
        \caption{Ablation study on diffusion timestep for encoding the measurement.}
        \label{fig:sdedit_fig_appdx}
     \end{minipage}
    \hfill
    \begin{minipage}{0.48\textwidth}
        \centering
        \includegraphics[width=\textwidth]{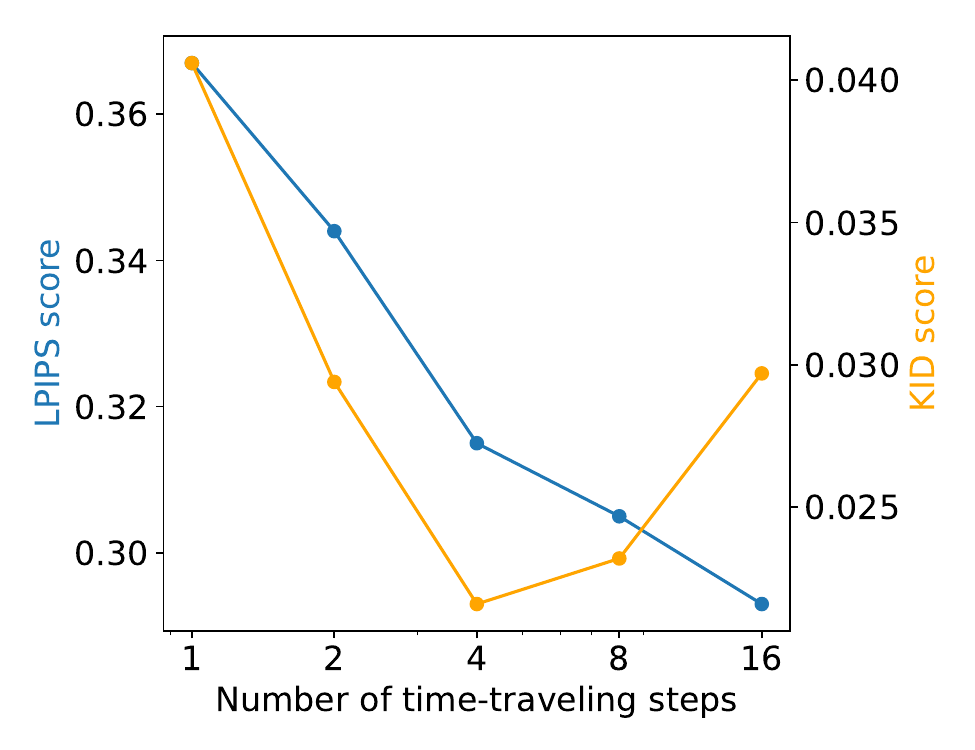}
        \caption{Ablation study on number of time-traveling steps.}
        \label{fig:tt_abl_fig_appdx}
      \end{minipage}
\end{figure}

\begin{table}[t]
\vspace{-5pt}
\small
\centering
\caption{Quantitative comparison between linear and deep operator architecture}
\label{tab:nn_op_linear_results}
\begin{tabular}{@{}ccccccc@{}}
\toprule
\multicolumn{1}{l}{}    & \multicolumn{3}{c}{AFHQ Motion Blur} \\ \midrule
Method & LPIPS $\downarrow$          & PSNR $\uparrow$          & KID $\downarrow$    \\ \hline
    Panl0           & 0.414 & 14.16 & 0.1590 \\
    PanDCP          & 0.371 & 17.63 & 0.1377 \\
    SelfDeblur      & 0.742 & 9.04 & 0.0650 \\
    MPRNet          & 0.324 & \underline{22.09} & 0.0382 \\
    DeblurGANv2     & 0.323 & \textbf{22.74} & 0.0350 \\
    BlindDPS        & 0.393 & 20.14 & 0.0913 \\
    GibbsDDRM       & 0.303 & 19.44 & 0.0265 \\
    GibbsDDRM*       & \underline{0.278} & 19.00 & 0.0180 \\ \midrule
    \textbf{\modelname{} (Kernel Operator)}   & \textbf{0.262}	& 21.20 & 	\textbf{0.0132} \\   
    \textbf{\modelname{} (U-Net Operator)}   & 0.343	& 18.93 & 	\underline{0.0146} \\
    \bottomrule
\end{tabular}
\vspace{-5pt}
\end{table}

\begin{figure}[t]
    \centering
    \includegraphics[width=\textwidth]{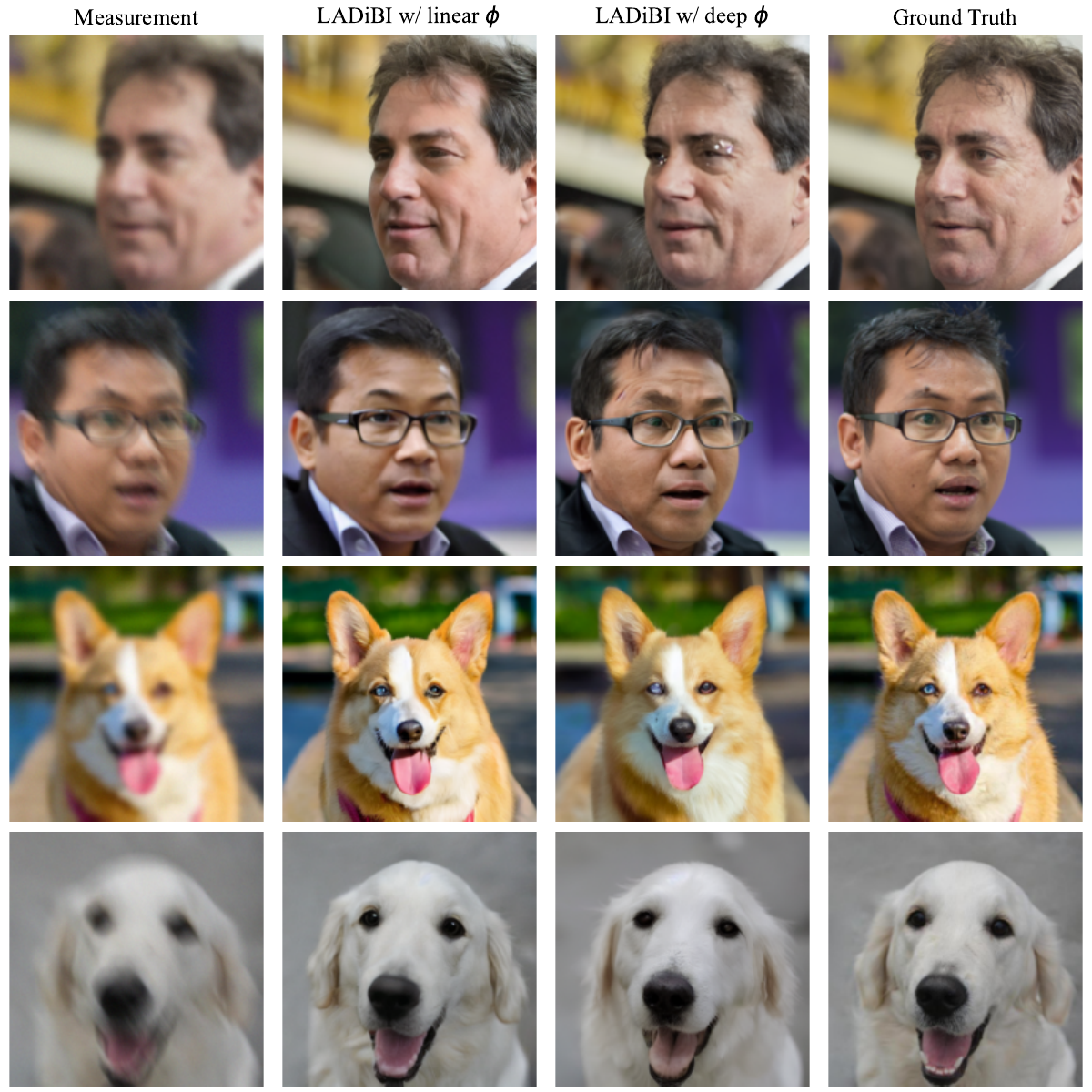}
    \caption{Comparison between linear and deep operator architecture on linear tasks.}
    \label{fig:nn_op_linear_fig_appdx}
\end{figure}

In this section, we provide additional ablation study on various hyperparameters and components of our \modelname{} algorithm.
All ablation studies are conducted on the motion deblurring task using the AFHQ dataset.

Figure \ref{fig:sdedit_fig_appdx} presents both the LPIPS and KID score for the value of the timestep $T_s$ at which we encode the measurement using SDEdit~\citep{sdedit}. Our experiments align with the observation in~\citet{sdedit}: If $T_s$ is too large most of the information about the measurement has been replaced by noise which does not allow the sampling process to leverage useful features of the degraded image. On the other hand, if $T_s$ is too small, the sampling process is not equipped with enough scheduled steps to reach the target distribution.

In addition, Figure \ref{fig:tt_abl_fig_appdx} presents the effectiveness of taking advantage of the time-traveling strategy. More time-traveling boosts the overall performance up to a specific value, after which we begin to notice a trade-off between perceptual clarity of the image and fidelity to the target distribution.

Finally, we evaluate the performance of our algorithm when using a unconstrained configuration for $\gA_{\hat{\phi}}$ in constrained tasks. In particular, we test the neural network architecture for the operator in linear inverse problems and we showcase quantitative results for the motion deblurring task in Table \ref{tab:nn_op_linear_results}, and qualitative samples in Figure \ref{fig:nn_op_linear_fig_appdx}. We observe that, although performing worse than \modelname{} employed with the aligned linear operator architecture, the neural operator is still capable of producing estimates of decent quality while also preserving a design that allows applicability to highly unconstrained image restoration task.
\clearpage
\section{Prior Coverage of Large Pre-trained Text-to-image Diffusion}
\label{sec:posterior}
\begin{figure}[h]
    \centering
    \begin{subfigure}{0.7\textwidth}
        \centering
    \includegraphics[width=\columnwidth, height=5.5cm]{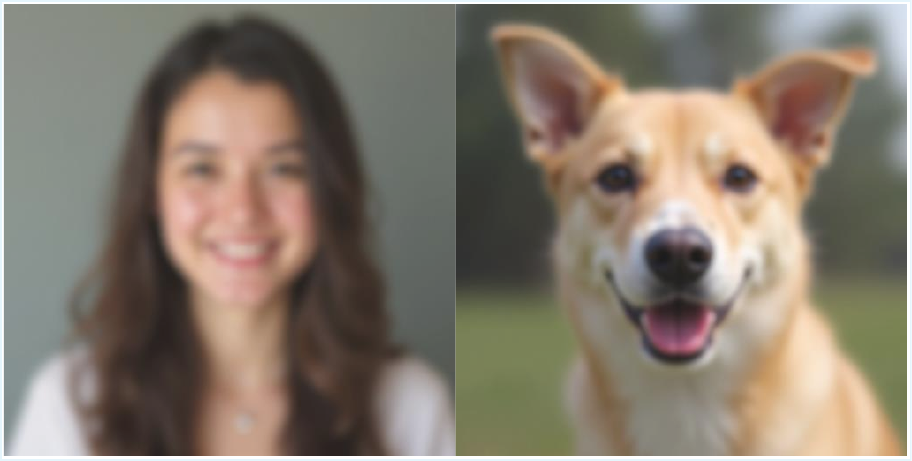}
    \caption{Generated images with Gaussian blur effect.}
    \label{fig: gaussian_blur_fig}
     \end{subfigure}
    \\
    \begin{subfigure}{0.7\textwidth}
        \centering
    \includegraphics[width=\columnwidth, height=5.5cm]{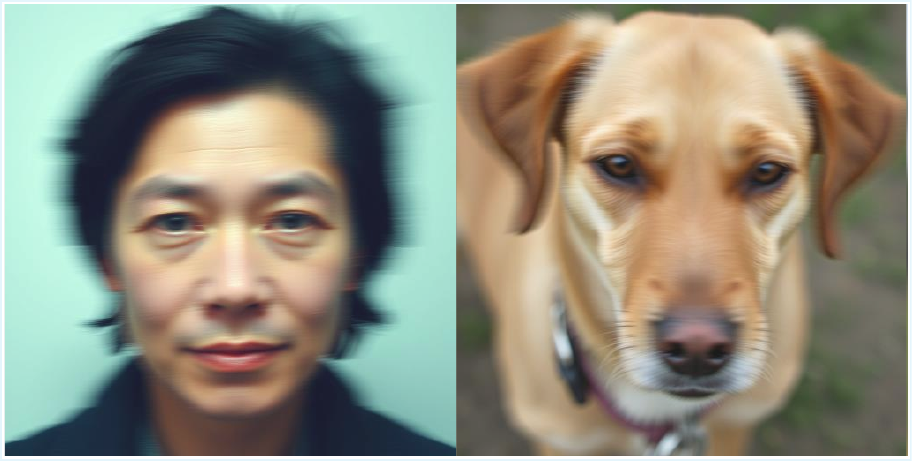}
    \caption{Generated images with motion blur effect.}
    \label{fig: motion_blur_fig}
      \end{subfigure}
    \\
    \begin{subfigure}{0.7\textwidth}
        \centering
    \includegraphics[width=\columnwidth, height=5.5cm]{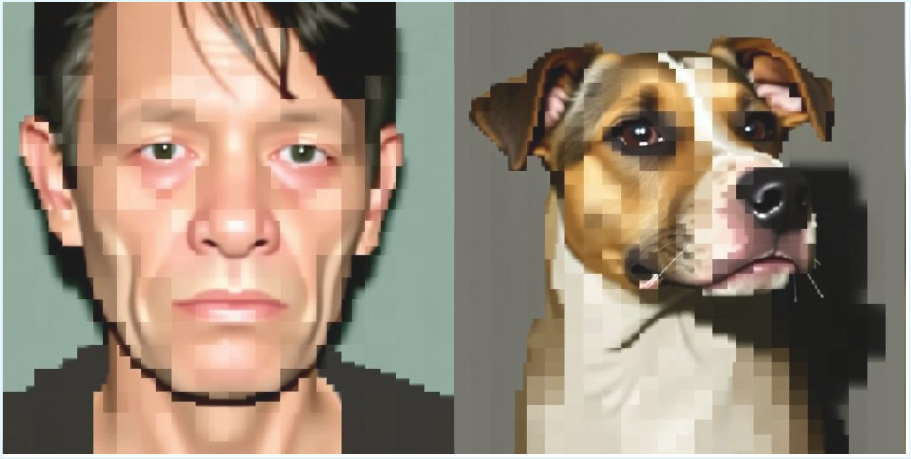}
    \caption{Generated images with JPEG compression effect.}
    \label{fig: motion_blur_fig}
      \end{subfigure}
    \caption{Drawn samples from the posterior distribution using our baseline model.}
    \label{fig:posterior}
\end{figure}

In this section we provide an empirical justification for using Stable Diffusion v1.4 as our base model.
To demonstrate this property, Figure \ref{fig:posterior} provides qualitative samples of images with Gaussian blur, motion blur and after JPEG compression. These examples show that Stable Diffusion 1.4 already capture rich distributions of both target images and common degradation artifacts.

\clearpage
\section{Limitations and Future Works}
\label{app:limitation}
In this section, we discuss the limitation of \modelname{} and potential future works to address these problems.

Similar to many training-free diffusion posterior sampling algorithms~\citep{blinddps,murata2023gibbsddrm,dps,he2024manifold}, our method is also sensitive to hyper-parameter tuning. We have provided details about our hyperparameter choices in the previous sections, and we will release our code in a public repository upon publication of this paper.

In addition, while generally simple, we do require the users to infer appropriate prompts from the measurement. An interesting future work direction can include automated prompt tuning similar to the method proposed in~\citet{chung2023prompt}.

Another notable drawback of our method is that \modelname{} requires significantly longer inference time in comparison to the best performing baseline (i.e. GibbsDDRM~\citep{murata2023gibbsddrm}) in order to obtain high quality restorations. With the general operator initialization, our algorithm can take around 5 minutes to complete on a single NVIDIA A6000 GPU while GibbsDDRM only takes around 30 seconds. Although this is justifiable by the larger optimization space that we operate on, investigating on how to reduce the inference time requirement is an interesting and critical next step for our work.

Lastly we would like to note that, while neural networks, as demonstrated in the previous sections, can serve as a general model family for various operator functional classes and achieve satisfactory results, obtaining state-of-the-art performance for linear tasks within a reasonable inference time still requires resorting to a linear kernel as the estimated operator. Exploring neural network architectures that can easily generalize across different operator functional classes while achieving state-of-the-art results efficiently remains an exciting direction for future work.

\section{Impact Statement}
\label{sec:ethic}
Lastly, since our algorithm leverages large pre-trained image generative models, we would like to address the ethical concerns, societal impact as well as the potential harm that can be caused by our method when being used inappropriately.

As a consequence of using large pre-trained text-to-image generative models, our method also inherits potential risks associated with these pre-trained models, including the propagation of biases, copyright infringement and the possibility of generating harmful content. We recognize the significance of these ethical challenges and are dedicated to responsible AI research practices that prevent reinforcing these ethical considerations. Upon the release our code, we are committed to implement and actively update safeguards in our public repository to ensure safer and more ethical content generation practices.

\end{document}